\documentclass[journal,onecolumn]{IEEEtran}

\makeatletter
\def\ps@headings{%
	\def\@oddhead{\mbox{}\scriptsize\rightmark \hfil \thepage}%
	\def\@evenhead{\scriptsize\thepage \hfil \leftmark\mbox{}}%
	\def\@oddfoot{}%
	\def\@evenfoot{}}
\makeatother 

\makeatletter
\def\endthebibliography{%
	\def\@noitemerr{\@latex@warning{Empty `thebibliography' environment}}%
	\endlist
}
\makeatother

\usepackage{graphicx}
\usepackage{amsmath, amsfonts,epsfig, multirow, floatflt}
\usepackage{amsthm}
\usepackage{amssymb}
\usepackage{nomencl}
\makenomenclature
\usepackage{etoolbox}
\usepackage{subfigure}
\usepackage{cite}
\usepackage{algorithm}
\usepackage{algorithmicx}
\usepackage{algpseudocode}
\usepackage{hyperref}
\usepackage{enumitem}
\usepackage{diagbox}
\usepackage{mathrsfs}
\usepackage{url}
\usepackage{color}

\usepackage{caption}
\usepackage{hhline}
\usepackage{array}
\usepackage{booktabs}
\usepackage{amsthm}
\usepackage{bm}
\usepackage[table,xcdraw]{xcolor}

\renewcommand\nomgroup[1]{%
	\item[\bfseries
	\ifstrequal{#1}{A}{System Model}{%
		\ifstrequal{#1}{B}{POMDP Model}{%
			\ifstrequal{#1}{C}{Other Symbols}{}
		}}%
	]
}
\allowdisplaybreaks[4]
\graphicspath{{figure/}}

\begin{document}

\title{Optimal Scheduling in IoT-Driven Smart Isolated Microgrids Based on Deep Reinforcement Learning}

\author{Jiaju Qi, Lei~Lei, {\it Senior Member, IEEE}, Kan~Zheng, {\it Senior Member, IEEE}, Simon X. Yang, {\it Senior Member, IEEE}, Xuemin (Sherman) Shen, {\it Fellow, IEEE}

\thanks{J. Qi, L. Lei, and S. X. Yang are with the School of Engineering, University of Guelph, Guelph, ON N1G 2W1, Canada (e-mail: jiaju@uoguelph.ca; leil@uoguelph.ca; syang@uoguelph.ca).}	
\thanks{K. Zheng is with the College of Electrical Engineering and Computer Sciences, Ningbo University, Ningbo, 315211, China. (email: zhengkan@nbu.edu.cn).}
\thanks{X. Shen is with the Department of Electrical and Computer Engineering, University of Waterloo, Waterloo, ON N2L 3G1, Canada (email: sshen@uwaterloo.ca).}
}

\maketitle

\begin{abstract}
	In this paper, we investigate the scheduling issue of diesel generators (DGs) in an Internet of Things (IoT)-Driven isolated microgrid (MG) by deep reinforcement learning (DRL). The renewable energy is fully exploited under the uncertainty of renewable generation and load demand. The DRL agent learns an optimal policy from history renewable and load data of previous days, where the policy can generate real-time decisions based on observations of past renewable and load data of previous hours collected by connected sensors. The goal is to reduce operating cost on the premise of ensuring supply-demand balance. In specific, a novel finite-horizon partial observable Markov decision process (POMDP) model is conceived considering the spinning reserve. In order to overcome the challenge of discrete-continuous hybrid action space due to the binary DG switching decision and continuous energy dispatch (ED) decision, a DRL algorithm, namely the hybrid action finite-horizon RDPG (HAFH-RDPG), is proposed. HAFH-RDPG seamlessly integrates two classical DRL algorithms, i.e., deep Q-network (DQN) and recurrent deterministic policy gradient (RDPG), based on a finite-horizon dynamic programming (DP) framework. Extensive experiments are performed with real-world data in an IoT-driven MG to evaluate the capability of the proposed algorithm in handling the uncertainty due to inter-hour and inter-day power fluctuation and to compare its performance with those of the benchmark algorithms.
\end{abstract}

\begin{IEEEkeywords}
Microgrid; Deep Reinforcement Learning; Energy Management
\end{IEEEkeywords}

\mbox{}
\nomenclature[A]{$D$,$d$}{The number/index of DGs}
\nomenclature[A]{$T$,$t$}{The number/index of time steps}
\nomenclature[A]{$B_{t}^{\mathrm{DG}_{d}}$}{The ON/OFF status of $\mathrm{DG}_{d}$}
\nomenclature[A]{$U_{t}^{\mathrm{DG}_{d}}$}{The binary switching decision of ${\mathrm{DG}_{d}}$}
\nomenclature[A]{$P_{t}^{\mathrm{DG}_{d}}$}{The output power of $\mathrm{DG}_{d}$}
\nomenclature[A]{$P_{\mathrm{min}}^{\mathrm{DG}}$,$P_{\mathrm{max}}^{\mathrm{DG}}$}{The minimum/maximum power limits}
\nomenclature[A]{$E_t$}{The SoC of the battery}
\nomenclature[A]{$E_{\mathrm{min}}$,$E_{\mathrm{max}}$}{The minimum/maximum energy level of the battery}
\nomenclature[A]{$\eta_{\mathrm{ch}}$,$\eta_{\mathrm{dis}}$}{The charging/discharging efficiencies of the battery}
\nomenclature[A]{$P_{t}^{\mathrm{E}}$}{The charging or discharging power of the battery}
\nomenclature[A]{$u_{t}$}{The charging or discharging status of the battery}
\nomenclature[A]{$P_{\mathrm{max}}^{\mathrm{E}}$}{The maximum charging or discharging power of the battery}
\nomenclature[A]{$P_{\mathrm{ch\_lim}}^{\mathrm{E}}$,$P_{\mathrm{dis\_lim}}^{\mathrm{E}}$}{The charging/discharging power limits of the battery}
\nomenclature[A]{$P_{t}^{\mathrm{PV}}$}{The output power of PV}
\nomenclature[A]{$P_{t}^{\mathrm{L}}$}{The load consumption power}
\nomenclature[A]{$P_{t}^{\mathrm{DG}_{d}\_{\mathrm{S}}}$}{A set-point for the DG output power }
\nomenclature[A]{$\delta_{t}$}{The surplus of generating capacity}
\nomenclature[A]{$P_{t}^{\mathrm{ERR}}$}{The unbalanced power after charging or discharging battery}
\nomenclature[A]{$P_{t}^{\mathrm{UB}}$}{The unbalanced power after DG output power correction}
\nomenclature[A]{$c_{t}^{\mathrm{DG}_{d}}$}{The DG power generation cost}
\nomenclature[A]{$c_{t}^{\mathrm{S\_DG}_{d}}$,$c_{t}^{\mathrm{R\_DG}_{d}}$,$c_{t}^{\mathrm{SR\_DG}_{d}}$}{The start-up/running/spinning reserve cost}
\nomenclature[A]{$c_{t}^{\mathrm{OP}}$}{The operating cost}
\nomenclature[B]{$S_{t}$}{The system state at time step $t$}
\nomenclature[A]{$P_{t}^{\mathrm{EL}}$}{The equivalent load}
\nomenclature[B]{$O_{t}$}{The observation at time step $t$}
\nomenclature[B]{$A_{t}$}{The action at time step $t$}
\nomenclature[B]{$A_{t}^{\mathrm{SW}}$,$A_{t}^{\mathrm{ED}}$}{The switching and ED action at time step $t$}
\nomenclature[B]{$\mathcal{A}$}{The action space}
\nomenclature[B]{$\mathcal{A}^{\mathrm{SW}}$,$\mathcal{A}^{\mathrm{ED}}$}{The switching and ED action space}
\nomenclature[B]{$\mathcal{A}_{k}^{\mathrm{ED}}$}{The ED action space corresponding to the switching action $k$}
\nomenclature[B]{$a_{k}^{\mathrm{SW}}$}{The $k$-th action in $\mathcal{A}^{\mathrm{SW}}$}
\nomenclature[B]{$a_{k}^{\mathrm{ED}}$}{An action in $\mathcal{A}_{k}^{\mathrm{ED}}$}
\nomenclature[B]{$k$}{The index of switching actions}
\nomenclature[B]{$H_{t}$}{The history at time step $t$}
\nomenclature[B]{$\tau$}{The length of history time window }
\nomenclature[B]{$\mathcal{H}$}{The history space}
\nomenclature[B]{$r(S_t,A_t)$}{The reward}
\nomenclature[B]{$c_{t}^{\mathrm{US}}$}{The cost of aggregated unserved or lost active power}
\nomenclature[B]{$C_1,C_2$}{The weights that indicate the relative importance of unserved power situation and lost power situation}
\nomenclature[B]{$\pi^*$}{The optimal policy}
\nomenclature[B]{$k^*$,${a_{k^*}^{\mathrm{ED}}}^*$}{The optimal switching and ED action}
\nomenclature[B]{$\hat{k}^*$,${a_{\hat{k}^*}^{\mathrm{ED}}}^*$}{The optimal switching and ED action derived by the proposed algorithm}
\nomenclature[B]{$\mu_k$,$\lambda_k$}{The actor and critic network for switching action $k$}
\nomenclature[B]{${a_{k}^{\mathrm{ED}}}^*$}{The optimal ED action of $\mathcal{A} _{k}^{\mathrm{ED}}$}
\nomenclature[B]{$\gamma$}{The discount factor}
\nomenclature[B]{$\theta$,$\omega$}{The weights of $\mu_k$ and $\lambda_k$}
\nomenclature[B]{$\alpha$,$\beta$}{The learning rate for updating $\theta$ or $\omega$}
\nomenclature[B]{$\hat{\mathcal{A}}^{\mathrm{SW}}$}{The simplified switching action space}
\nomenclature[B]{$m$}{A simplified switching action}
\nomenclature[B]{$m^*$}{The optimal simplified switching action}
\nomenclature[B]{${\mathcal{A}}_{m}^\mathrm{SW}$}{A subset within ${\mathcal{A}}^\mathrm{SW}$ that corresponds to $m$}
\nomenclature[B]{$k_m$}{The unique switching action in ${\mathcal{A}}_{m}^\mathrm{SW}$ that is mapped to $m$}
\nomenclature[A]{$a_d$,$b_d$,$c_d$}{The quadratic/monomial/constant coefficient of DG power generation cost curve}
\nomenclature[A]{$C_{\mathrm{R}}$,$C_{\mathrm{S}}$,$C_{\mathrm{SR}}$}{The fixed running/start-up/spinning reserve cost coefficient}
\nomenclature[B]{$R^{\pi}$}{The cumulative reward}
\printnomenclature
\nomenclature[A]{$\mathrm{DG}_d$}{The $d$-th DG}

	\section{Introduction}
    Microgrids (MGs) are small-scale low- or medium-voltage distribution networks comprising various on-site generators. Nowadays, Internet of Things (IoT) plays an essential role in MGs \cite{9069178}. In a smart MG driven by the IoT technologies, the system operator (SO) is deployed in the edge or cloud server and connects with networked sensors embedded in various MG components \cite{8642832}. The real-time data collected by these sensors are transmitted to the SO through the IoT communication networks to reflect the status of the MG system \cite{9631955}. By fully analyzing the sensory data, the SO makes intelligent energy management decisions to drive the smart MG devices, such as diesel generators (DGs) \cite{8458217}.
    
    MGs can operate either interconnected or isolated from the main distribution grid \cite{9069178}. Isolated MGs are deployed in many remote areas all over the world \cite{butt2021recent}, which often rely on a base load generation source, such as reciprocating DGs of heavy fuel oil. As DGs incur high costs of electricity in isolated MGs due to the cost of fuel transportation and delivery, renewable energy sources (RES) such as wind and solar are a clean complementary power source that greatly reduces the operating costs \cite{8281479}. In this paper, we focus on the optimal scheduling problem in smart isolated MGs, which aims at minimizing the operating cost by fully exploiting the RES power, while satisfying the critical requirement to maintain balance between power generation and consumption. The main challenge stems from the variability and uncertainty in RES generation and load demand, making it difficult to guarantee energy balance with efficient operations of DGs.\par 
    
    Compared with connected MGs, isolated MGs are less flexible in dealing with the above challenge as they do not have power support from the main grid and usually lack advanced control hardware for demand side management. Fortunately, the energy storage elements in MGs, such as electro-chemical battery, can charge or discharge to compensate for short-term unbalance, and thus help level off the power fluctuations to some extent. Moreover, the battery can act as an additional power source to minimize operating cost of DGs. As the maximum amount of energy that can be charged or discharged at a certain point of time is limited by the state of charge (SoC) of the battery, which is in turn determined by the previous charging/discharging behavior, optimal scheduling becomes a sequential decision problem for a dynamic system where earlier decisions influence future available choices. 
    
    In this paper, we leverage deep reinforcement learning (DRL) to learn an optimal scheduling policy from history RES and load data of previous days where the policy can generate real-time decisions based on observations of past RES and load data of previous hours. The main advantage of such an approach is that it does not require prediction models and therefore is not susceptible to prediction errors.\par

    \subsection{Related Works}
   Energy management for MGs include several typical functions, i.e., optimal scheduling (such as energy storage management, energy dispatch (ED), unit commitment (UC) \cite{azaroual2019optimal}, etc.), demand response \cite{ccimen2020microgrid}, and energy trading \cite{VANLEEUWEN2020114613}. Existing works in this area can be classified into two broad categories, depending on whether the MG is connected \cite{arcos2019review,chen2022robust} or isolated \cite{sachs2016two,Lara2019}. Our study falls into optimal scheduling in isolated MGs, which has been extensive studied in the literature. To determine the daily energy scheduling operation, various mathematical optimization models have been proposed, where the objectives are normally to minimize the operating cost considering various constraints such as operational, energy balance, and reserve constraints. Different problem formulations have been studied, such as mixed integer linear programming (MILP)\cite{Luna2017,Moretti2019,Conte2019}, mixed integer quadratic programming (MIQP)\cite{Solanki2017,Solanki2019}, mixed integer and integer-free second-order cone programming (SOCP)\cite{Zia2019,Giraldo2019}. In order to address the challenge of uncertainty in RES generation and load demand, several approaches such as stochastic programming (SP)\cite{Olivares2015}, model predictive control (MPC)\cite{sachs2016two,Solanki2019,Violante2020a}, robust optimization\cite{Giraldo2019,Lara2019}, and chance-constraint programming (CCP) \cite{Li2019,Li2022a} have been adopted. All of the above methods require forecast models or probabilistic models for the unknown data, where robustness to model inaccuracy is limited. \par

     To avoid the above limitations, intelligent energy management policies can be derived by leveraging reinforcement learning (RL) techniques. Compared with the non-RL methods, RL algorithms can learn the optimal control policies by trial and error based on real-world data, and do not require rigorous mathematical models for RES and load \cite{o2018uncertainty}. In other words, RL methods can directly learn optimal policies from history data, without the need of prior modeling or parameter identification. DRL is a promising branch of RL, where the powerful deep neural networks (DNNs) make it possible to learn and fit complex patterns better. Moreover, it is easy for DRL methods to solve large-scale optimization problems in real-time after training the DNNs, since only the forward propagation in the networks is involved. \par
     
     In recent years, DRL has been adopted for energy management of MGs with the aim for learning optimal policies of demand response \cite{hu2022multi,mocanu2018line,ruelens2014demand}, energy trading \cite{xiao2017energy,kim2014dynamic,xiao2018reinforcement}, as well as optimal scheduling in both connected MGs \cite{en12122291,shuai2020online,Du2020} and isolated MGs \cite{franccois2016deep,9277511}. In \cite{en12122291}, Ji et al. adopted a classical model-free value-based DRL algorithm namely deep Q-network (DQN) \cite{Mnih2015a} to achieve real-time scheduling of a connected MG, which did not require an explicit model or a predictor for unknown data. The real-time dispatch of DGs and battery as well as power transactions with the main grid are optimized based on this approach. Considering a similar online scheduling problem for a connected MG, Huang et al. adopted a model-based DRL algorithm, namely MuZero, to perform online optimization of MG under uncertainties \cite{shuai2020online}, which combined a Monte-Carlo tree search method with the neural network model. Du et al. considered a scenario where multiple MGs are connected to the main grid, where each MG adopts a model-free Monte-Carlo DRL algorithm to learn its retail pricing policy \cite{Du2020}. While the above works focus on connected MGs, DRL-based solutions for isolated MGs have received less attention. Franois et al. presented a value-based DRL algorithm to efficiently operate the storage devices in an isolated MG\cite{franccois2016deep}. However, only three simple discrete actions are considered, i.e., charge, discharge, and idle. In \cite{9277511}, Lei et al. proposed new DRL algorithms to solve the energy dispatch problem in an isolated MG. The algorithms combined classical actor-critic algorithms namely deep deterministic policy gradient (DDPG) \cite{Lillicrap2015} and recurrent deterministic policy gradient (RDPG) \cite{Heess2015} with dynamic programming approach, and could effectively address the instability problem, finite-horizon setting and partial observable problem. However, the startup and shutdown of DGs are not considered, which results in inefficient operation. 

 Based on the review of related works, some important research gaps in the field are identified as follows. 
 
 \begin{itemize}
 	\item In non-RL methods, spinning reserve is usually considered a deterministic \cite{sachs2016two,Solanki2017,Solanki2019} or probabilistic constraint \cite{Li2019,Li2022a} to guarantee energy balance under power fluctuations. Deterministic constraint generally leads to inefficient DG operation \cite{Li2019}, while probabilistic constraint is vulnerable to model or parameter deviations.
 	\item When applying DRL for optimal scheduling of MGs, one important challenge is the discrete-continuous hybrid action space induced by the binary DG switching decision and the continuous ED decision. Classical DRL algorithms are not suitable to handle the hybrid action space. Previous works in DRL usually approximate the hybrid space by discretization \cite{en12122291,shuai2020online}, which suffers from performance loss.  
 \end{itemize}

    \subsection{Contributions} 
    
   The main contributions are summarized as follows. 
   \begin{enumerate}
   	\item  A novel partial observable Markov decision process (POMDP) model is conceived to learn the optimal scheduling policy that can provide sufficient spinning reserve capacity to maintain energy balance while minimizing the operating cost in the IoT-driven isolated MG. Instead of considering spinning reserve as constraints, a penalty for energy imbalance is included in the reward function, and the recalculation of new set-points during real-time operation is captured in the system transition function, so that the DRL agent can learn to set the optimal amount of spinning reserve. 
   	\item  A novel DRL algorithm, i.e., \emph{hybrid action finite-horizon RDPG (HAFH-RDPG)}, is proposed. The integration of actor-critic RDPG and value-based DQN characteristics enables our algorithm to have great advantages in dealing with the discrete-continuous hybrid action space. Moreover, the DRL algorithm is embedded within the finite-horizon value iteration framework to improve the convergence stability and performance.  
   	
   	\item The capability of DRL in handling uncertainty is evaluated at two time-scales of power fluctuation: inter-hour and inter-day. The capability to deal with inter-hour fluctuation is assessed by comparing the performance of the proposed algorithm in POMDP and Markov decision process (MDP) environments, respectively. The capability to deal with inter-day fluctuation is assessed by comparing the performance of HAFH-RDPG algorithm when it is trained using same-day and previous-days data, respectively. Moreover, we analyze whether the learned policy can make efficient use of battery in achieving energy shifting and providing sufficient spinning reserve with minimum cost.
   \end{enumerate}  
	The remainder of the paper is organized as follows. The IoT-driven isolated MG system model is introduced in Section II. The POMDP model is developed in Section III. The design of HAFH-RDPG algorithm is presented in Section IV. In section V, experiments are conducted to verify the efficiency of HAFH-RDPG algorithm. Finally, the conclusion is drawn in Section VI.

	\section{System Model}
	
\begin{figure}[t]
		\centering
		\includegraphics[width=0.45\textwidth]{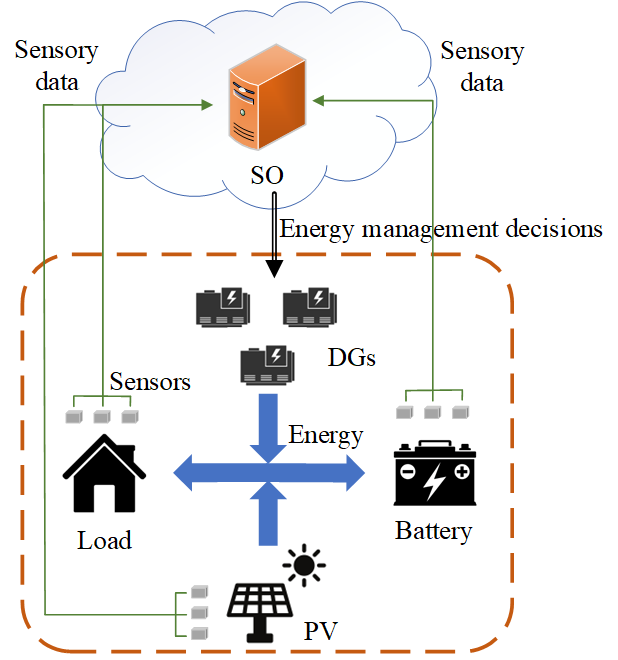}
		\caption{The schematic diagram of the IoT-driven smart isolated MG system.}
		\label{MGsystem}
	\end{figure}

	As shown in Fig. \ref{MGsystem}, the IoT-driven smart isolated MG system considered in this paper can be divided into several parts, i.e., the controllable generation units (i.e., DGs), uncontrollable generation units (i.e., photovoltaic cell (PV)), an energy storage device (i.e., battery storage), loads, and a SO (i.e., DRL-based intelligent energy management system) to perform energy scheduling. To reflect the status of the smart MG system, sensors are deployed to collect real-time data on the load power consumption, the output power generated by the PV panels, and the SoC of the battery storage. The sensory data is delivered to the SO in the edge or cloud server, where the DRL agent of the SO processes the data to make intelligent decisions on optimal scheduling of DGs. The information of sensory data and control decisions are transmitted between the SO in edge/cloud server and the smart devices in MG through the IoT communication networks.

	
	The intra-day operation of our MG system model is divided into $T$ time steps, indexed by $\{1,\cdots,T\}$. The interval of each time step is denoted as $\Delta t$.\par
	
    \subsection{Diesel Generator Model}
     Consider there are $D$ DGs in the MG, i.e., $\mathrm{DG}_{1},\cdots,\mathrm{DG}_{D}$, where $D > 1$. Let $B_{t}^{\mathrm{DG}_{d}}$ denote the ON/OFF status of $\mathrm{DG}_{d}$, $\forall d\in\{1,\cdots,D\}$, \textit{at the beginning} of time step $t$, where $1$ stands for ON and $0$ for OFF. 
 
   Let $U_{t}^{\mathrm{DG}_{d}}$ denote the binary switching decision of ${\mathrm{DG}_{d}}$ at time step $t$. After $U_{t}^{\mathrm{DG}_{d}}$ is determined at time step $t$, the ON/OFF status of $\mathrm{\mathrm{DG}}_{d}$ will be set accordingly for the rest of the time step. Note that the ON/OFF status $B_{t}^{\mathrm{DG}_{d}}$ at the beginning of time step $t$ is the same as the switching decision $U_{t-1}^{\mathrm{DG}_{d}}$ at the previous time step $t-1$. Therefore, we have 
  \begin{equation}
  	\label{eq2}
  	U_{t-1}^{\mathrm{DG}_{d}} = B_{t}^{\mathrm{DG}_{d}}.
  \end{equation}
     
     At time step $t$, let $P_{t}^{\mathrm{DG}_{d}}$ denote the output power of $\mathrm{DG}_{d}$, $\forall d\in\{1,\cdots,D\}$. When a DG is switched on, its output power can be any real number between the minimum and maximum power limits, i.e., $P_{\mathrm{min}}^{\mathrm{DG}}$ and $P_{\mathrm{max}}^{\mathrm{DG}}$. Thus, we have \par 
     \begin{equation}
     	\label{eq1.5}
     	\begin{array}{ll}
     		P_{t}^{\mathrm{DG}_d}=0,  & \mathrm{if} \ U_{t}^{\mathrm{DG}_{d}}=0\\
     		P_{\min}^{\mathrm{DG}}\leqslant P_{t}^{\mathrm{DG}_d}\leqslant P_{\max}^{\mathrm{DG}}  & \mathrm{if} \ U_{t}^{\mathrm{DG}_{d}}=1\\
     	\end{array}.
     \end{equation}	



    \subsection{Battery Model}
    
    Let $E_{t}$ be SoC of the battery at time step $t$, which is constrained by the minimum and maximum energy levels as
    \begin{equation}
    	\label{eq3}
    E_{\min}\leqslant E_t\leqslant E_{\max}.
    \end{equation}
    
    The SoC of battery $E_{t+1}$ at time step $t+1$ can be calculated based on the SoC state $E_{t}$ at time step $t$ as
    \begin{equation}
    E_{t+1}=E_{t}+\eta_{\mathrm{ch}} u_{t} P_{t}^{\mathrm{E}} \Delta t-(1-u_{t})P_{t}^{\mathrm{E}}\Delta t/\eta_{\mathrm{dis}},
    \label{eq4}
    \end{equation}
    \noindent where $\eta_{\mathrm{ch}}$ and $\eta_{\mathrm{dis}}$ denote the charging and discharging efficiencies of the battery, respectively. $P_{t}^{\mathrm{E}}$ indicates the charging or discharging power of the battery, and $u_{t}$ denotes charging or discharging status, which is 0 if battery is discharging and
    1 otherwise. \par
    
    Let $P_{\mathrm{max}}^{\mathrm{E}}$ be the maximum charging or discharging power of the battery. The charging and discharging power limits of the battery can be calculated separately as
    \begin{equation}
    P_{\mathrm{ch\_lim}}^{\mathrm{E}}=\min(P_{\mathrm{max}}^{\mathrm{E}},(E_{\mathrm{max}}-E_{t})/(\eta_{\mathrm{ch}}\Delta t)),
    \end{equation} 
    \noindent and
    \begin{equation}
    P_{\mathrm{dis\_lim}}^{\mathrm{E}}=\min(P_{\mathrm{max}}^{\mathrm{E}},\eta_{\mathrm{dis}}(E_{t}-E_{\mathrm{min}})/\Delta t).   
    \end{equation} 
    
     \subsection{Energy Balance and Spinning Reserve}
    At each time step $t$, the total amount of energy dispatched by the DGs, PV, and battery should meet the load demand whenever possible, i.e.,
      \begin{equation}
      \label{balance}
\sum_{d=1}^{D}P_{t}^{\mathrm{DG}_{d}}+P_{t}^{\mathrm{PV}}+\eta_{\mathrm{ch}} u_{t} P_{t}^{\mathrm{E}}-(1-u_{t})P_{t}^{\mathrm{E}}/\eta_{\mathrm{dis}}=P_{t}^{\mathrm{L}}
  \end{equation}   
	\noindent where $P_{t}^{\mathrm{PV}}$ is the output power of PV and $P_{t}^{\mathrm{L}}$ is the load consumption at time step $t$. \par    
   
    At the beginning of each time step $t$, a set-point for the DG output power $P_{t}^{\mathrm{DG}_{d}\_{\mathrm{S}}}$ is determined by the DRL agent. During real-time operation within the time step, as the actual PV generation $P_{t}^{\mathrm{PV}}$ and load demand $P_{t}^{\mathrm{L}}$ are observed, corrective actions must be taken to ensure the energy balance. For an isolated MG without power support from the main grid, the spinning reserve is an important resource to compensate for power shortages due to uncertainty in PV generation and load demand. Both the DGs and battery can provide spinning reserve, where the battery is given a priority due to its lower cost and faster response than the DGs. Similarly, the battery is leveraged to cope with power oversupply before adjusting DG output power.  \par
    \subsubsection{Charging/discharging of battery}
    In order to maintain energy balance in \eqref{balance}, the battery needs to follow load, where the charging/discharging power $P_{t}^{\mathrm{E}}$ and charging/discharging status $u_{t}$ can be determined as follows. First, a variable $\delta_{t}$ is defined to represent the surplus of generating capacity at time step $t$ assuming the DG output power is set to $P_{t}^{\mathrm{DG}_{d}\_{\mathrm{S}}}$ , i.e.,
    \begin{equation}
	\delta_{t}=\sum_{d=1}^{D}P_{t}^{\mathrm{DG}_{d}\_{\mathrm{S}}}+P_{t}^{\mathrm{PV}}-P_{t}^{\mathrm{L}}.
	\label{delta}
	\end{equation}
When a surplus is generated, i.e., $\delta _t>0$, the excess power is stored in the battery and the battery is in the charging status. On the contrary, when the power generation of PV and DGs is insufficient, i.e., $\delta _t<0$, the battery is in the discharging status to provide the lacked power. 
	In this way, $u_{t}$ can be determined based on $\delta _t$ as 
	\begin{equation}
		u_t= \left\{
		\begin{array}{ll}
			\text{0, } & \mathrm{if} \ \delta _t<0\\
			\text{1, } & \mathrm{otherwise} \\
		\end{array}\right. ,
	\label{ut}
	\end{equation}

	\noindent and $P_{t}^{\mathrm{E}}$ can be derived based on $\delta _t$ as
	\begin{equation}	
	P_{t}^{\mathrm{E}}=\left\{
	\begin{array}{ll}
		\min \left( -\delta _t, P_{\mathrm{dis\_lim}}^{\mathrm{E}} \right) , & \mathrm{if} \ \delta _t<0\\
		\min \left( \delta _t, P_{\mathrm{ch\_lim}}^{\mathrm{E}} \right) , & \mathrm{otherwise} \\
	\end{array}\right. .
\label{eq10}
	\end{equation}
	
	To elaborate on \eqref{eq10}, when $\delta_{t}<-P_{\mathrm{dis\_lim}}^{\mathrm{E}}$, even discharging the battery cannot provide enough power to meet the load demand. When $\delta_{t}>P_{\mathrm{ch\_lim}}^{\mathrm{E}}$, the excessive power is beyond the charging capability of the battery. In the above cases, we will need to adjust the output power of DGs. For ease of notation, we denote the unbalanced power after charging/discharging battery as
		\begin{align}	  
	P_{t}^{\mathrm{ERR}}=\left\{
	\begin{array}{ll}
\delta_{t}+P_{\mathrm{dis\_lim}}^{\mathrm{E}}, & \mathrm{if} \ \delta_{t}<-P_{\mathrm{dis\_lim}}^{\mathrm{E}} \\
\delta_{t}-P_{\mathrm{ch\_lim}}^{\mathrm{E}}, & \mathrm{if} \ \delta_{t}>P_{\mathrm{ch\_lim}}^{\mathrm{E}}\\
0, & \mathrm{otherwise} \\
	\end{array}\right. .
	\end{align}
	
%

	
	  \subsubsection{Adjustment of DG output power}
	To maintain energy balance in \eqref{balance}, the total output power of DGs is adjusted subject to the power limits as follows:
	
		\begin{align}
		\label{dgadjust}		  
	&\sum_{d=1}^{D}P_{t}^{\mathrm{DG}_{d}}= \IEEEnonumber \\
	&\left\{
	\begin{array}{ll}
	\max\Big\{\sum_{d=1}^{D}U_{t}^{\mathrm{DG}_{d}}P_{\min}^{\mathrm{DG}},P_{t}^{\mathrm{Adjusted}}\Big\} & \mathrm{if} \ P_{t}^{\mathrm{ERR}}>0 \\	
	\min\Big\{\sum_{d=1}^{D}U_{t}^{\mathrm{DG}_{d}}P_{\max}^{\mathrm{DG}},P_{t}^{\mathrm{Adjusted}}\Big\} & \mathrm{if} \ P_{t}^{\mathrm{ERR}}<0 \\  
		\end{array}\right. ,
	\end{align}
	\noindent where
\begin{equation}
	P_{t}^{\mathrm{Adjusted}}=\sum_{d=1}^{D}P_{t}^{\mathrm{DG}_{d}\_{\mathrm{S}}}-P_{t}^{\mathrm{ERR}}.
\end{equation}

      We define the unbalanced power after DG output power correction as 
	
	      \begin{equation}
	\label{UB}
P_{t}^{\mathrm{UB}}=\sum_{d=1}^{D}P_{t}^{\mathrm{DG}_{d}}+P_{t}^{\mathrm{PV}}+\eta_{\mathrm{ch}} u_{t} P_{t}^{\mathrm{E}}-(1-u_{t})P_{t}^{\mathrm{E}}/\eta_{\mathrm{dis}}-P_{t}^{\mathrm{L}}
	\end{equation}   
	When $P_{t}^{\mathrm{UB}}=0$, energy balance in \eqref{balance} is achieved. If $P_{t}^{\mathrm{UB}}>0$, the power generation is larger than demand, and the excessive power will go to load bank, which will be lost. Large $P_{t}^{\mathrm{UB}}$ beyond the maximum capacity of load bank will cause the undesirable reverse power flow in the MG. If $P_{t}^{\mathrm{UB}}<0$, the generation is smaller than demand, which will result in some of the loads unserved.  \par
	
	\subsection{Operating Cost}	
The operating cost includes four parts. For any $\mathrm{DG}_{d}$, $\forall d\in\{1,\cdots,D\}$, the DG power generation cost $c_{t}^{\mathrm{DG}_{d}}$ can be derived by the conventional quadratic cost function in \eqref{costDG}. The start-up cost $c_{t}^{\mathrm{S\_DG}_{d}}$ in \eqref{costst} and the running cost $c_{t}^{\mathrm{R\_DG}_{d}}$ in \eqref{costru} are penalty costs associated with turning on $\mathrm{DG}_{d}$ and the ON status of $\mathrm{DG}_{d}$, respectively. Finally, the spinning reserve cost $c_{t}^{\mathrm{SR\_DG}_{d}}$ in \eqref{costSR} depends on the spinning reserve capacity, i.e., the amount of unused capacity, provided by $\mathrm{DG}_{d}$. 
    \begin{equation}
	\label{costDG}
	c_{t}^{\mathrm{DG}_{d}}=[a_{d}(P_{t}^{\mathrm{DG}_{d}})^{2}+b_{d}P_{t}^{\mathrm{DG}_{d}}+c_{d}]\Delta t, \forall d\in\{1,\cdots,D\},
	\end{equation} 
		\begin{equation}
	\label{costst}
	c_{t}^{\mathrm{S\_DG}_{d}}=C_{\mathrm{S}}U_{t}^{\mathrm{DG}_{d}}(1-B_{t}^{\mathrm{DG}_{d}})\Delta t, \forall d\in\{1,\cdots,D\},
	\end{equation} 
		\begin{equation}
	\label{costru}
	c_{t}^{\mathrm{R\_DG}_{d}}=C_{\mathrm{R}}U_{t}^{\mathrm{DG}_{d}}\Delta t, \forall d\in\{1,\cdots,D\},
	\end{equation}
		\begin{equation}
	\label{costSR}
	c_{t}^{\mathrm{SR\_DG}_{d}}=C_{\mathrm{SR}}(U_{t}^{\mathrm{DG}_{d}}P_{\max}^{\mathrm{DG}} -P_{t}^{\mathrm{DG}_{d}})\Delta t.
	\end{equation}  
	\noindent Note that in \eqref{costDG}-\eqref{costSR}, $a_{d}$, $b_{d}$, $c_{d}$, $C_{\mathrm{S}}$, $C_{\mathrm{R}}$, and $C_{\mathrm{SR}}$ are the coefficients, where $a_{d}$, $b_{d}$, and $c_{d}$ are fitted to fuel consumption curves for DGs; while $C_{\mathrm{S}}$, $C_{\mathrm{R}}$, and $C_{\mathrm{SR}}$ are weights that indicate the relative importance of the start-up cost, the running cost, and the spinning reserve cost, respectively.  \par
	
	Therefore, the operating cost $c_{t}^{\mathrm{OP}}$ is the sum of the above costs for all the DGs, i.e.,		 
	\begin{align}
	\label{opcost}
	c_{t}^{\mathrm{OP}}= \sum_{d=1}^{D}\left(U_{t}^{\mathrm{DG}_{d}}c_{t}^{\mathrm{DG}_{d}}+c_{t}^{\mathrm{S\_DG}_{d}}+c_{t}^{\mathrm{R\_DG}_{d}}+c_{t}^{\mathrm{SR\_DG}_{d}}\right).
	\end{align}

	Note that \eqref{dgadjust} only provides the total DG output power after adjustment. The output power of individual DGs can be derived by minimizing the operating cost $c_{t}^{\mathrm{OP}}$ in \eqref{opcost} after the total output power is determined.\par

	\section{POMDP Model}
	In this section, we  formulate a finite-horizon POMDP model based on which the optimal scheduling decisions using DRL can be made. The objective of our POMDP model is to minimize the operating cost and ensure the supply-demand balance, where the uncertainty of both fluctuating loads and stochastic generation of PV are taken into account. \par

    \subsection{State and Observation}

  Let $S_{t}=\left(P_{t}^{\mathrm{EL}}, E_t,\{B_{t}^{\mathrm{DG}_{d}}\}_{d=1}^{D}\right)$ be the system state at time step $t$, $\forall t\in\{1,\cdots,T\}$, where $P_{t}^{\mathrm{EL}}=P_{t}^{\mathrm{L}}-P_{t}^{\mathrm{PV}}$ is the equivalent load. Due to the uncertainties of load consumption and PV generation, the agent is unable to observe $P_{t}^{\mathrm{EL}}$ at the beginning of time step $t$, but receives observation $O_{t}=(P_{t-1}^{\mathrm{EL}},E_{t},\{B_{t}^{\mathrm{DG}_{d}}\}_{d=1}^{D})$ instead. Note that $P_{t-1}^{\mathrm{EL}}=P_{t-1}^{\mathrm{L}}-P_{t-1}^{\mathrm{PV}}$ is the equivalent load at time step $t-1$.


    \subsection{Action}
    Let $A_t=\left( A_{t}^{\mathrm{SW}}, A_{t}^{\mathrm{ED}} \right)$ be the action at time step $ t\in\{1,\cdots,T\}$, where $A_{t}^{\mathrm{SW}}=\{U_{t}^{\mathrm{DG}_d}\}_{d=1}^{D}$ denotes the DG switching action, which is a vector of \emph{discrete} variables. $A_{t}^{\mathrm{ED}}=\{P_{t}^{\mathrm{DG}_{d}\_{\mathrm{S}}}\}_{d=1}^{D}$ represents the ED action that determines the DG output power set-point, which is a vector of \emph{continuous} variables.\par

    The action space $\mathcal{A}=\mathcal{A}^{\mathrm{SW}} \times \mathcal{A}^{\mathrm{ED}}$. As $\mathcal{A}^{\mathrm{SW}}$ is $D$ permutations of $0$ and $1$, its size is $2^D$.  Let $a_{k}^{\mathrm{SW}}$ denote the $k$-th action in the switching action space $\mathcal{A}^{\mathrm{SW}}$, where $k\in \left\{ 1,...,2^D \right\}$ is the index of the switching action. Since there is a one-to-one correspondence between $a_{k}^{\mathrm{SW}}$ and $k$, we substitute $k$ for $a_{k}^{\mathrm{SW}}$ for convenience in the rest of this paper.\par
    
%
    According to \eqref{eq1.5}, $\mathcal{A}^{\mathrm{ED}}$ is related to the switching action $k$. Therefore, let $\mathcal{A}_{k}^{\mathrm{ED}}$ denote the ED action space corresponding to the chosen switching action $k$. Let $a_{k}^{\mathrm{ED}}$ denote an action belonging to the ED action space $\mathcal{A}_{k}^{\mathrm{ED}}$, i.e., $a_{k}^{\mathrm{ED}}\in \mathcal{A} _{k}^{\mathrm{ED}}
    $.
    
    
    
    \subsection{History}
    As only the observation $O_{t}$ is available instead of state $S_{t}$, the agent is provided with history $H_{t}$ to derive the action $A_{t}$. The history is tailored for the optimal scheduling of isolated MG and defined as $H_{t}=(P_{t-\tau}^{\mathrm{EL}},P_{t-\tau+1}^{\mathrm{EL}},\cdots,P_{t-1}^{\mathrm{EL}},E_{t},\{B_{t}^{\mathrm{DG}_{d}}\}_{d=1}^{D})$, $\forall t\in\{1,\cdots,T\}$, where $\tau$ is the length of time window to look into the past. Note that $H_{t}$ includes history data of equivalent load statistics $\{P_{t'}^{\mathrm{EL}}\}_{t'=t-\tau}^{t-1}$ to implicitly predict the equivalent load state $P_{t}^{\mathrm{EL}}$ at time step $t$. Let $\mathcal{H}$ denote the history space. \par
    
    
    \subsection{Reward Function}
    The optimization objective in the IoT-driven MG system is to minimize the operating cost within the considered time horizon, i.e., one day, while guaranteeing energy balance. Therefore, we define the reward function as 
   \begin{align}
    \label{rt}
    r(S_t,A_t)= -(c_{t}^{\mathrm{OP}}+c_{t}^{\mathrm{US}}),
    \end{align}
    \noindent where $c_{t}^{\mathrm{OP}}$ is the operating cost in \eqref{opcost}.
  $c_{t}^{\mathrm{US}}$ represents the cost of aggregated unserved or lost active power, i.e.,
    \begin{equation}
    	\label{costUS}
    	c_{t}^{\mathrm{US}}=\left\{
    	\begin{array}{ll}
    		C_{1}P_{t}^{\mathrm{UB}}\Delta t, & \mathrm{if} \ P_{t}^{\mathrm{UB}}>0\\
    		-C_{2}P_{t}^{\mathrm{UB}}\Delta t, & \mathrm{if} \ P_{t}^{\mathrm{UB}}<0 \\
    		0, & \mathrm{otherwise} \\
    	\end{array}\right. ,
    \end{equation}
    \noindent where $C_1$ and $C_2$ are weights that indicate the relative importance of unserved power situation and lost power situation, and $P_{t}^{\mathrm{UB}}$ can be derived according to \eqref{UB}.\par 
 
    \subsection{Transition Probability}
    The state transition probability is derived as
    \begin{equation}
    	\begin{split}
    		\mathrm{Pr}\left( S_{t+1}|S_t,A_t \right) =\mathrm{Pr}\left( P_{t+1}^{\mathrm{L}}|P_{t}^{\mathrm{L}} \right) \mathrm{Pr}\left( P_{t+1}^{\mathrm{PV}}|P_{t}^{\mathrm{PV}} \right) \\
    		\mathrm{Pr}\left( E_{t+1}|E_t,P_{t}^{\mathrm{L}},P_{t}^{\mathrm{PV}},A_{t}^{\mathrm{ED}} \right) \mathrm{Pr}\left( B_{t+1}|A_{t}^{\mathrm{SW}} \right) ,
    	\end{split}
        \label{pr}
    \end{equation}
    \noindent where the transition probabilities of load demands $\mathrm{Pr}\left( P_{t+1}^{\mathrm{L}}|P_{t}^{\mathrm{L}} \right)$ and PV power outputs $\mathrm{Pr}\left( P_{t+1}^{\mathrm{PV}}|P_{t}^{\mathrm{PV}} \right)$ are not available, but samples of the trajectory can be obtained from real-world data. The transition probability of SoC $\mathrm{Pr}\left( E_{t+1}|E_t,P_{t}^{\mathrm{L}},P_{t}^{\mathrm{PV}},A_{t}^{\mathrm{ED}} \right)$ can be calculated through \eqref{eq4}, \eqref{delta}, \eqref{ut}, and \eqref{eq10}. The last transition probability $\mathrm{Pr}\left( B_{t+1}|A_{t}^{\mathrm{SW}} \right)$ is for the switching state, which can be easily inferred from \eqref{eq2}. 
    
    Similarly, the history transition probability is derived as
    \begin{equation}
    	\begin{split}
    		\mathrm{Pr}\left( H_{t+1}|H_t,A_t \right) =\mathrm{Pr}\left( P_{t}^{\mathrm{L}}|P_{t-\tau}^{\mathrm{L}},\dots,P_{t-1}^{\mathrm{L}} \right)  \\
    		\mathrm{Pr}\left( P_{t}^{\mathrm{PV}}|P_{t-\tau}^{\mathrm{PV}},\dots,P_{t-1}^{\mathrm{PV}} \right) \mathrm{Pr}\left( B_{t+1}|A_{t}^{\mathrm{SW}} \right) \\
    		\mathrm{Pr}\left( E_{t+1}|E_t,P_{t}^{\mathrm{L}},P_{t}^{\mathrm{PV}},A_{t}^{\mathrm{ED}} \right),
    	\end{split}
    	\label{prHt}
    \end{equation}
    \noindent where the transition probabilities of load demands $\mathrm{Pr}\left( P_{t}^{\mathrm{L}}|P_{t-\tau}^{\mathrm{L}},\dots,P_{t-1}^{\mathrm{L}} \right)$ and PV power outputs $\mathrm{Pr}\left( P_{t}^{\mathrm{PV}}|P_{t-\tau}^{\mathrm{PV}},\dots,P_{t-1}^{\mathrm{PV}} \right)$ can be obtained by the samples of the trajectory from real-world data. The transition probability of SoC $\mathrm{Pr}\left( E_{t+1}|E_t,P_{t}^{\mathrm{L}},P_{t}^{\mathrm{PV}},A_{t}^{\mathrm{ED}} \right)$ and the transition probability of switching state $\mathrm{Pr}\left( B_{t+1}|A_{t}^{\mathrm{SW}} \right)$ are the same as the transition probabilities in \eqref{pr}.
    

 \subsection{Optimization Objective}

   We define the cumulative reward $R^{\pi}$ as the sum of rewards over the finite horizon under a given policy $\pi$, i.e.,
    \begin{equation}
        \label{cumulative reward}
        R^{\pi}=\sum_{t=1}^T{r\left( S_t,A_t \right)}.
    \end{equation}
Our objective is to derive the optimal policy $\pi^*$ to maximize the expected cumulative reward, i.e.,
\begin{equation}
    \label{objective1}
    \pi ^*=\underset{\pi}{\mathrm{arg}\max}\mathbb{E}\left[ R^{\pi} \right]. 
\end{equation}

	\section{DRL Solution}
\begin{figure}[t]
		\centering
		\includegraphics[width=0.45\textwidth]{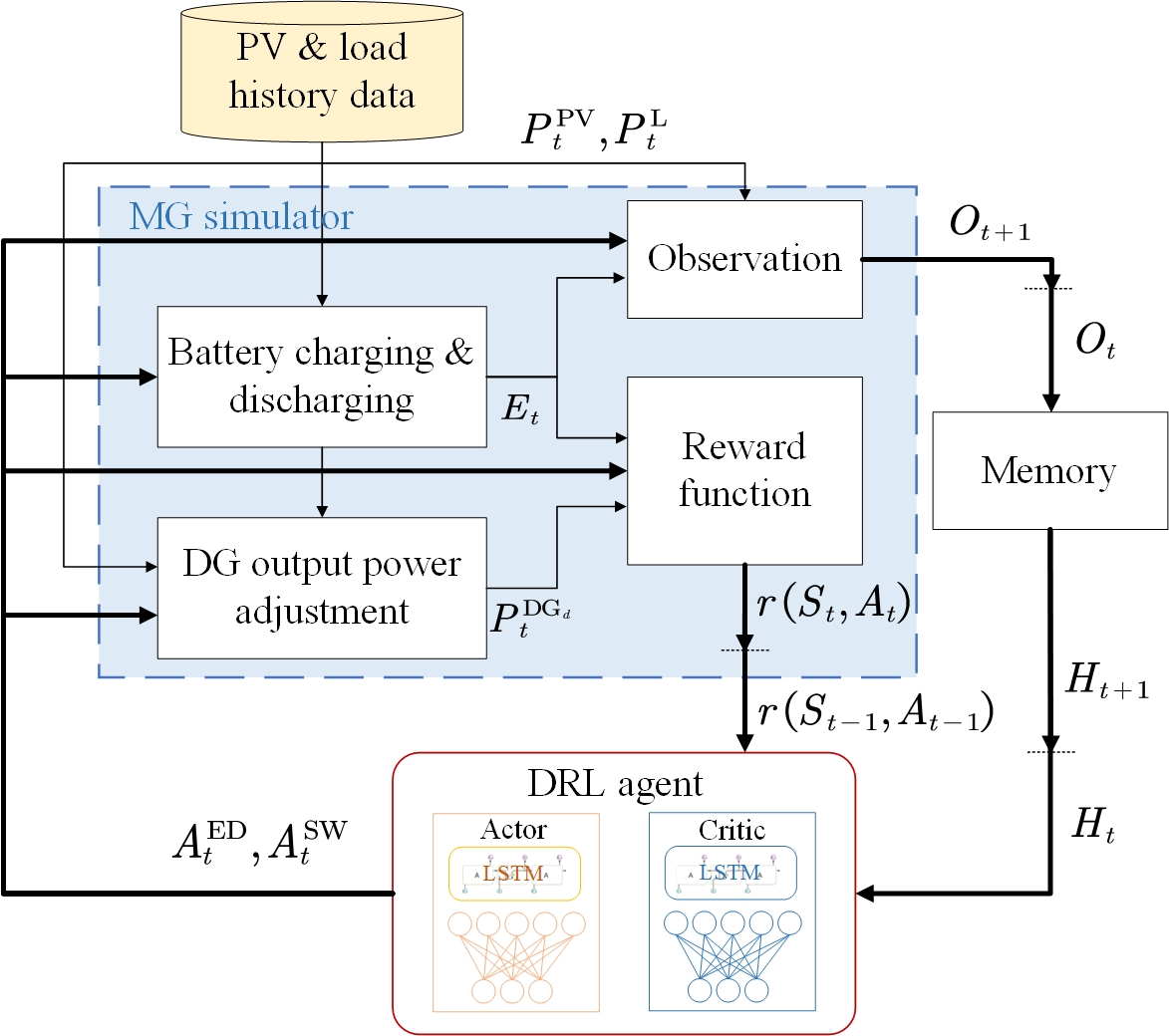}
		\caption{The schematic description of the learning process of the DRL agent.}
		\label{flow}
	\end{figure}

	In this section, we present a DRL algorithm to solve the POMDP model formulated in Section III. A schematic description of the learning process is provided in Fig. \ref{flow}. The DRL agent is trained using history PV and load data and can be used for real-time optimal scheduling after training. \par
	
	\subsection{Dealing with the hybrid action space}
    Since the action $A_t$ in our POMDP model includes both continuous and discrete variables, the action space $\mathcal{A}$ is a discrete-continuous hybrid space, while most of the existing DRL algorithms require either a discrete space or a continuous space. Inspired by the P-DQN algorithm in \cite{xiong2018parametrized}, we break the optimization problem down into two steps and design a new algorithm to solve this problem.

	Firstly, the optimization objective in \eqref{objective1} is equivalent to finding the policy $\pi^*$ that can maximize the action value $Q\left( H_t,k,a_{k}^{\mathrm{ED}}\right)$, i.e., 
	\begin{equation}
		\pi^*(H_t)=\left( k^*,{a_{k^*}^{\mathrm{ED}}}^*\right) =\underset{k\in \left\{ 1,...,2^D \right\}, a_{k}^{\mathrm{ED}}\in \mathcal{A} _{k}^{\mathrm{ED}}}{\mathrm{arg}\max}Q\left( H_t,k,a_{k}^{\mathrm{ED}} \right) ,
  \label{objective}
	\end{equation}
    \noindent where $k^*$ and ${a_{k^*}^{\mathrm{ED}}}^*$ denote the optimal switching action and ED action, respectively.
			
	For the discrete switching action $k$, there are $2^D$ choices. For any given switching action $k\in\left\{1,...,2^D\right\}$, we are able to select the best ED action ${a_{k}^{\mathrm{ED}}}^{*}$ from the ED action space $\mathcal{A}_{k}^{\mathrm{ED}}$ by solving
	\begin{equation}
		\label{akED*}	
	{a_{k}^{\mathrm{ED}}}^*=\underset{a_{k}^{\mathrm{ED}}\in \mathcal{A} _{k}^{\mathrm{ED}}}{\mathrm{arg}\max}Q\left( H_{t},k,a_{k}^{\mathrm{ED}} \right), \forall k\in \left\{1,...,2^D\right\} .
    \end{equation}
	However, as $a_{k}^{\mathrm{ED}}$ is a continuous variable, \eqref{akED*} cannot be directly solved by classical value-based algorithms such as Q-learning and DQN \cite{xiong2018parametrized} unless the action space is discretized, which will result in performance loss due to discretization error. In contrast, the actor-critic algorithms can efficiently handle the continuous action space. Therefore, our algorithm adopts the actor-critic framework to calculate the optimal ED action ${a_{k}^{\mathrm{ED}}}^*$.
	
	Specifically, for a given switching action $k, \ \forall k\in \left\{ 1,...,2^D \right\}$, an actor network $\mu_k \left( \cdot ;\theta \right) :\mathcal{H} \rightarrow \mathcal{A} _{k}^{\mathrm{ED}}$ is introduced to approximate the optimal ED action ${a_{k}^{\mathrm{ED}}}^{*}$ from the continuous space $\mathcal{A} _{k}^{\mathrm{ED}}$ such that
	\begin{equation}
 \label{akED*2}	
		{a_{k}^{\mathrm{ED}}}^*\approx \mu _k\left( H_t;\theta \right) ,
    \end{equation} 
	\noindent where $\theta$ is the weights of $\mu_k$. Correspondingly, a critic network $\lambda_k \left( \cdot ;w \right)$ is applied to evaluate the ED action ${a}_{k}^{\mathrm{ED}^*}$ and approximate the action value $Q\left( H_{t},k,a_{k}^{\mathrm{ED}^*} \right)$, i.e.,
	\begin{equation}
 \label{lambdak}
		Q\left( H_{t},k,a_{k}^{\mathrm{ED}^*} \right) \approx \lambda_k \left( H_{t},{a}_{k}^{\mathrm{ED}^*};w \right) ,
	\end{equation}	
	\noindent where $w$ denotes the weights of $\lambda_k$. 
	
	Once the optimal ED action ${a_{k}^{\mathrm{ED}}}^*$ for all the switching action $k, \forall k\in \left\{ 1,...,2^D \right\}$, are derived, we could choose the optimal switching action $\hat{k}^*$ from the $2^D$-dimensional discrete switching action space. The above action value $Q\left( H_{t},k,a_{k}^{\mathrm{ED}^*} \right)$ given by the critic network $\lambda_k$ can be used to evaluate the switching action $k$ when the corresponding optimal ED action ${a_{k}^{\mathrm{ED}}}^*$ is applied. We choose the switching action $k$ with the largest action value as the optimal solution, i.e.,
	\begin{equation}
		\label{k*}
		\hat{k}^*=\underset{k\in \left\{ 1,...,2^D \right\}}{\mathrm{arg}\max}Q\left( H_{t},k,{a_{k}^{\mathrm{ED}}}^* \right) .
	\end{equation}
	
\begin{figure}[t]
	\centering
	\includegraphics[width=0.4\textwidth]{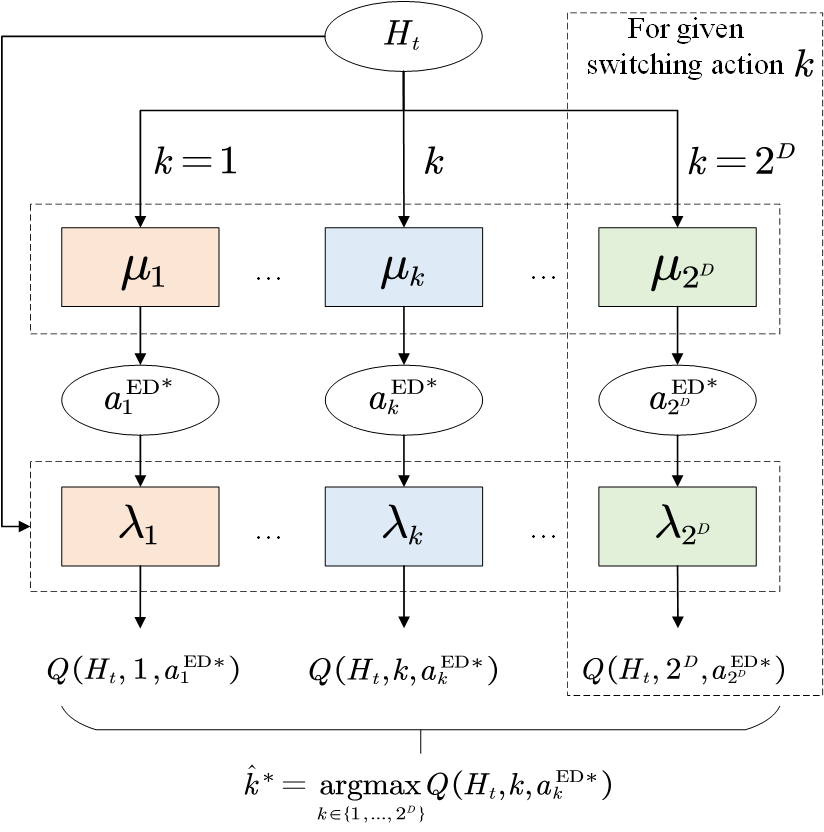}
	\caption{The neural network framework of HAFH-RDPG.}
	\label{structure}
\end{figure}

	
	As shown in Fig. \ref{structure}, the above two-step framework is summarized as follows: 
\begin{enumerate}
    \item  For each switching actions $k\in\left\{1,...,2^D\right\}$, select the corresponding optimal ED action ${a_{k}^{\mathrm{ED}}}^{*}$ by \eqref{akED*} in the continuous action space.
    \item Derive the optimal switching action $\hat{k}^*$ in the discrete action space according to \eqref{k*}.
\end{enumerate}
 
     In the following, we present Theorem 1 that states the optimal policy for the objective in \eqref{objective} can be derived by the proposed two-step framework. 
	
	
	\newtheorem{thm}{\bf Theorem}
	\begin{thm}
		\label{thm_optimal}
       For any history $H_t$, the action $\left( \hat{k}^*,{a_{\hat{k}^*}^{\mathrm{ED}}}^* \right)$ derived by the two-step framework is as good as the optimal action $\pi^*(H_t)$ for the objective in \eqref{objective}, i.e.,
        \begin{equation}
        	\label{thmeq0}
	Q\left(H_t,\hat{k}^*,{a_{\hat{k}^*}^{\mathrm{ED}}}^* \right)=Q\left(H_t, k^*,{a_{k^*}^{\mathrm{ED}}}^* \right) .
        \end{equation}
        
	\end{thm}

   The proof of Theorem 1 is given in Appendix A.

 In order to train the actor and critic networks, we focus on the action value function $Q\left( H_t,A_{t} \right) =Q\left( H_t,A_{t}^{\mathrm{SW}},A_{t}^{\mathrm{ED}} \right)$ based on the Bellman equation as 
	\begin{equation}
		\begin{split}
			Q\left( H_t,A_{t}^{\mathrm{SW}},A_{t}^{\mathrm{ED}} \right) = \underset{S_t|H_t}{\mathbb{E}}\left[r(S_t,A_{t}^{\mathrm{SW}},A_{t}^{\mathrm{ED}})\right] \\
			+\underset{H_{t+1}|H_t,A_{t}^{\mathrm{SW}},A_{t}^{\mathrm{ED}}}{\mathbb{E}}\left[\gamma\cdot \underset{k\in \left\{ 1,...,2^D \right\}}{\max}\underset{a_{k}^{\mathrm{ED}}\in \mathcal{A} _{k}^{\mathrm{ED}}}{\max}Q\left( H_{t+1},k,a_{k}^{\mathrm{ED}} \right) \right]  ,
		\end{split}
		\label{bellman}
	\end{equation}

\noindent where $\gamma$ is the discount factor that satisfies $0< \gamma \leqslant 1$. 

    Thus, the loss function $L_k(w)$ of the critic network $\lambda_k$, $\forall k\in \left\{ 1,...,2^D \right\} $ can be calculated as the mean square error, i.e.,
    \begin{equation}
    	\begin{split}
	L_k(w )&=\underset{H_t\sim\rho^\mu_k,a_{k}^{\mathrm{ED}}\sim\mu_k}{\mathbb{E}}\left[ \left( y_{t,k}-\lambda _k\left( H_t,a_{k}^{\mathrm{ED}};w \right) \right) ^2 \right] ,
	    \end{split}
    \end{equation} 
    \noindent where $\rho^\mu_k$ denotes the discounted state visitation distribution following policy $\mu_k$. $y_{t,k}$ denotes the target for each critic network $\lambda_k$ at time step $t$, which is defined based on the Bellman equation in \eqref{bellman} as
\begin{equation}
	\begin{split}
		y_{t,k}
		&\approx r_t+\gamma \underset{k\in \left\{ 1,...,2^D \right\}}{\max}\lambda _k\left( H_{t+1},\mu _k\left( H_{t+1};\theta \right) ;w \right) .
	\end{split}
\end{equation}
    
    By minimizing the loss function $L_k(w)$ via stochastic gradient descent, the weight $w$ of $\lambda_k$ can be updated at each time step by
	\begin{equation}
		\begin{split}
 w \gets &w +2\beta \left( y_{t,k}-\lambda _k\left( H_t,a_{k}^{\mathrm{ED}};w \right) \right) \nabla _{w}\lambda _k\left( H_t,a_{k}^{\mathrm{ED}};w \right) ,
        \end{split}
    \end{equation}
    \noindent where $\beta$ is the learning rate for updating $w$.
    
    The loss function of actor network $\mu_k$, $\forall k\in \left\{ 1,...,2^D \right\}$ can be defined as
    \begin{equation}
    	\begin{split}
    	L_k\left( \theta \right)
    	&\approx -\underset{H_0\sim \rho ^0}{\mathbb{E}}\left[ \lambda _k\left( H_0,\mu _k\left( H_t;\theta \right) ;w \right) \right], 
    	\end{split}
    \end{equation}
\noindent where $\rho_0$ is the distribution of the initial history $H_0$.
    

    We use stochastic gradient decent to sample deterministic policy gradient \cite{silver2014deterministic}, and the weight $\theta$ of $\mu_k$ is updated by
    \begin{equation}
    \theta \gets \theta -\alpha \nabla _{\theta}\mu _k\left( H_t;\theta \right) \nabla _{{a_{k}^{\mathrm{ED}}}}\lambda _k\left( H_t,{a_{k}^{\mathrm{ED}}};w \right) |_{{a_{k}^{\mathrm{ED}}}=\mu _k\left( H_t;\theta \right)},
    \end{equation}
    \noindent where $\alpha$ is the learning rate for updating $\theta$.

	
	
	\subsection{Designing the framework}
	
	As we focus on energy management within one day with $T$ time steps, our task corresponds to a finite horizon POMDP model. There are a few popular DRL algorithms for solving POMDP such as Deep Recurrent Q-Learning (DRQN) \cite{Hausknecht2015} and RDPG. The basic ideas of DRQN and RDPG are to add recurrency to DQN and DDPG algorithms, respectively, by replacing the first fully-connected layer with a recurrent long short-term memory (LSTM) layer. However, the above DRL algorithms are developed for an infinite horizon setting, while the value function, i.e., critic, and the policy, i.e., actor, are normally dependent on the time steps for the finite horizon case. Therefore, we propose a novel DRL algorithm named HAFH-RDPG, which adopts a framework that is a combination of dynamic programming and DRL, where RDPG with fixed target is embedded within the framework of finite-horizon value iteration. \par

	
	In HAFH-RDPG, the finite-horizon value iteration starts from the time step $T$, and uses backward induction to iteratively derive the value function and optimal policy for each time step $t\in \left\{ T,T-1,...,1 \right\} $, until it reaches the first time step $t=1$. In each time step, the RDPG algorithm is used to solve a simple one-step POMDP where the target actor and critic networks, i.e., $\lambda _{t}^{'}$ and $\mu _{t}^{'}$ are fixed to be the trained actor and critic networks of the next time step, i.e., $\lambda _{t+1}$ and $\mu_{t+1}$, which greatly increases stability and performance. \par 
	
	
	Algorithm \ref{alg-HAFH-DDPG} describes the pseudocode of HAFH-RDPG. The calculation of the hybrid actions and the method to update networks are presented by two functions separately, i.e., Algorithms \ref{alg-HAFH-DDPG-funcaction} and \ref{alg-HAFH-DDPG-train}. In the Action Calculation function, we introduce multiple actor networks to calculate the optimal ED actions for every switching action. According to \eqref{k*}, we choose the optimal switching action whose critic network has the largest output value and the corresponding ED action. In the Networks Updating function, in order to update the critic networks, the minibatch of transitions $\mathcal{B}$ are further divided into non-overlapping subsets according to the switching actions. The critic network $\lambda_k$ is updated by a minibatch of transitions $\mathcal{J}$ in which the switching action of each transition is $k$. After that, the actor networks can be updated based on $\mathcal{B}$ using the sampled policy gradient.

 The proposed HAFH-RDPG algorithm works for MG systems with arbitrary number of DGs. When the number of DGs changes, we only need to add or remove actor-critic pairs correspondingly and the training process will be in the same way as before.

	\begin{algorithm}[t]
		\caption{HAFH-RDPG algorithm}
		\label{alg-HAFH-DDPG}
		\begin{algorithmic}[1]  
			\State Randomly initialize actor networks $\mu_k \left( h ;\theta_k \right)$  and critic networks $\lambda_k \left( h,a ;w_k \right)$ with weights $\theta_k=\theta_k^0$ and $w_k=w_k ^0$, respectively, where $\forall k\in \left\{ 1,...,2^D \right\}$. Initialize target networks $\mu _{k}^{'}\left( h ;\theta_k^{'} \right)$ and $\lambda_{k}^{'}\left( h,a ;w_k^{'} \right)$ with $\theta_k ^{'}\gets \theta_k $ and $w_k ^{'}\gets w_k $;
			\For{$t = T, ..., 1$}
			\State Initialize replay buffer $\mathcal{R}$ 
			\State Initialize a random process $\mathcal{N}$ for action exploration
			\For{episode $e = 1, ..., M$}
			\State Receive history $H_t^{\left(e\right)}$
			\State ${k_{t}^{\left( e \right)}},{A_{t}^{\mathrm{ED}}}^{\left( e \right)}$=\Call{Action}{$H_t^{\left(e\right)}$,$\mu_k\left( h ;\theta_k \right)$,$\lambda_k\left( h,a ;w_k \right)$}
			\State Execute action and observe reward $r_t^{(e)}$ and next observation $O_{t+1}^{(e)}$
			\State Store transition $\left( H_{t}^{\left( e \right)},k_{t}^{\left( e \right)},{A_{t}^{\mathrm{ED}}}^{\left( e \right)},r_{t}^{\left( e \right)},O_{t+1}^{\left( e \right)} \right) $ into $\mathcal{R}$
			\State Sample a random minibatch of $N$ transitions $\mathcal{B}=\left( H_{t}^{\left( i \right)},k_{t}^{\left( i \right)},{A_{t}^{\mathrm{ED}}}^{\left( i \right)},r_{t}^{\left( i \right)},O_{t+1}^{\left( i \right)} \right) $ from $\mathcal{R}$
			\If{$t=T$}
			\State Set the target $y_{T}^{\left( i \right)}=r_{T}^{\left( i \right)}$
			\Else
			\State Construct $H_{t+1}^{\left( i \right)}$
			\State Set the target $$y_{t}^{\left( i \right)}=r_{t}^{\left( i \right)}+\gamma\cdot\underset{k\in \left\{ 1,...,2^D \right\}}{\max} \lambda _{k}^{'}\left( H_{t+1}^{\left( i \right)},\mu _{k}^{'}\left( H_{t+1}^{\left( i \right)};\theta_k \right);w_k \right) $$
			\EndIf
	        \State \Call{Update}{$\mathcal{B}$,$y_{t}^{\left( i \right)}$,$\mu_k\left( h ;\theta_k \right)$,$\lambda_k\left( h,a ;w_k \right)$}
			\EndFor 
			\State Update the target networks: $w_k^{'} \gets w_k$, $\theta_k^{'} \gets \theta_k$
			\State Save weight of the actor networks: $ \theta_{k,t} \gets \theta_k$
			\State Reset weight of the actor and critic networks to initial value: $\theta_k \gets \theta_k^0$, $w_k \gets w_k^0$
			\EndFor 
			\end{algorithmic}  
			\end{algorithm} 
	
	\begin{algorithm}[t]  
		\caption{Action calculation}
		\label{alg-HAFH-DDPG-funcaction}
		\begin{algorithmic}[1]  
			\Function{Action}{$H_t^{\left(e\right)}$,\ $\mu_k\left( h ;\theta_k \right)$,\ $\lambda_k\left( h,a ;w_k \right)$}
			\For{switching action $k = 1, ..., 2^D$ }
			\State Calculate ED action ${a_{k}^{\mathrm{ED}}}^{*\left( e \right)}=\mu _k\left( H_{t}^{\left( e \right)};\theta_k \right) $
			\EndFor
			\State Select switching action $ k_{t}^{\left( e \right)}=\underset{k\in \left\{ 1,...,2^D \right\}}{\mathrm{arg}\max}\lambda _k\left( H_{t}^{\left( e \right)},{a_{k}^{\mathrm{ED}}}^{*\left( e \right)};w_k \right)  $ 
			\State Select ED action $ {A_{t}^{\mathrm{ED}}}^{\left( e \right)}={a_{k_t^{(e)}}^{\mathrm{ED}}}^{*\left( e \right)}$
			\State Adjust switching action $k_{t}^{\left( e \right)}$ based on $\epsilon $-greedy policy
			\State Adjust ED action ${A_{t}^{\mathrm{ED}}}^{\left( e \right)}$ with exploration noise
			\State \Return{${k_{t}^{\left( e \right)}}$ and ${A_{t}^{\mathrm{ED}}}^{\left( e \right)}$}
			\EndFunction
		\end{algorithmic}  
	\end{algorithm} 

    \begin{algorithm}[t]  
	    \caption{Networks updating}
	    \label{alg-HAFH-DDPG-train}
	    \begin{algorithmic}[1]  
		    \Function{Update}{$\mathcal{B}$,\ $y_{t}^{\left( i \right)}$,\ $\mu_k\left( h ;\theta_k \right)$,\ $\lambda_k\left( h,a ;w_k \right)$}
		    \For{$k = 1, ..., 2^D$ }
		    \State Sample the minibatch of transitions $\mathcal{J} = \left( H_{t}^{\left( j \right)},k_{t}^{\left( j \right)},{A_{t}^{\mathrm{ED}}}^{\left( j \right)},r_{t}^{\left( j \right)},O_{t+1}^{\left( j \right)} \ |\  k_{t}^{\left( j \right)}=k \right) $ from $\mathcal{B}$ and corresponding targets $y_{t}^{\left( j \right)}$
		    \State Update the critic networks by minimizing the loss:
		    $$
		    L_k=\frac{1}{\left| \mathcal{J} \right|}\sum_{j \in \mathcal{J} }{\left( y_{t}^{\left( j \right)}-\lambda _k\left( H_{t}^{\left( j \right)},{A_{t}^{\mathrm{ED}}}^{\left( j \right)};w_k \right) \right)}^2,$$
		    $$w_k\gets w_k+\beta \nabla _{w_k}L_k	$$
		    \EndFor
		    \State Update the actor networks using the sampled policy gradient:
		    $$
		    \begin{aligned}
		    \nabla _{\theta_k}J_k\approx\frac{1}{\left| \mathcal{B} \right|} \sum_{i}{\left[ \nabla _a\lambda _k\left( h,a;w_k \right) |_{h=H_{t}^{\left( i \right)},a=\mu _k\left( H_{t}^{\left( i \right)};\theta _k \right)} \right.}\\
		    	{\left. \nabla _{\theta _k}\mu _k\left( H_{t}^{\left( i \right)};\theta _k \right)\right]}  , \theta_k \gets \theta_k +\alpha \nabla _{\theta_k}J_k.
		    \end{aligned}
		    $$
		    
		    \EndFunction
	    \end{algorithmic}  
    \end{algorithm} 
	
	\subsection{Reducing the algorithm complexity}

	Since the proposed method is a DRL-based algorithm, it updates the network parameters instead of the Q values of all state-action pairs in every iteration. The computational complexity of training each pair of actor-critic networks in HAFH-RDPG is the same with that of the classical RDPG algorithm, which is associated with the size of neural networks instead of the state and action spaces. In specific, the per-iteration computational complexity is mainly dominated by the matrix multiplication in the LSTM layer and fully-connected layers. However, note that the HAFH-RDPG algorithm requires $\left| {\mathcal{A}}^{\mathrm{SW}} \right|=2^D$ pairs of actor and critic networks, thus the computational complexity is $O(2^D)$. As the number of DGs $D$ increases, the number of actor and critic networks and thus the computational complexity of HAFH-RDPG increases exponentially. \par
 
  In order to reduce the computational complexity, we map the action space to a new space with reduced dimensionality. For simplicity of explanation, we consider the parameters of all $D$ DGs are identical in the following. When the parameters of only a subset of DGs are identical, our method can be applied to the subset of DGs instead of all the DGs. The new switching action space $\hat{\mathcal{A}}^{\mathrm{SW}}$ is defined as $\hat{\mathcal{A}}^{\mathrm{SW}}=\left\{ 0,1,...,D \right\}$, where a simplified switching action in the new space $m\in\hat{\mathcal{A}}^{\mathrm{SW}}$ corresponds to the number of DGs that are ON. Compared with $\mathcal{A}^{\mathrm{SW}}$, the size of $\hat{\mathcal{A}}^{\mathrm{SW}}$ is reduced from $2^D$ to $D+1$, which can in turn reduce the number of neural networks and the computation complexity from $O(2^D)$ to $O(D)$.
	
	Note that there is a one-to-many relationship between $m$ and $k$. Specifically, each $m$ corresponds to $C_{D}^{m}$ possible switching actions $k$. When $m$ DGs should be ON, the corresponding switching actions $k$ form a subset ${\mathcal{A}}_{m}^\mathrm{SW}$ within ${\mathcal{A}}^\mathrm{SW}$. In other words, ${\mathcal{A}}^\mathrm{SW}$ can be divided into $D+1$ non-overlapping subsets, i.e.,
	\begin{equation}
		\label{cup}
		\bigcup_{m=0}^D{\mathcal{A} _{m}^{\mathrm{SW}}}=\mathcal{A} ^{\mathrm{SW}}.
	\end{equation}
	
	
    In the following, we introduce a strategy that maps a simplified switching action $m\in\hat{\mathcal{A}}^{SW}$ to a unique switching action $k_m\in \mathcal{A} _{m}^{\mathrm{SW}}$.
    
	
	
	\textbf{DG Selection Strategy}. First, a priority value is assigned for each $\mathrm{DG}_d, \forall d\in\{1,2,...,D\}$, so that $\mathrm{DG}_1$ has the highest priority and $\mathrm{DG}_D$ has the lowest. If $m$ DGs should be ON at time step $t$ where $\forall m\in \left\{ 0,1,...,D \right\}$, the switching action under this strategy $k_m$ should be
		\begin{equation}
			\label{atUCstr}
		k_m = \left( \left\{ U_{t}^{\mathrm{DG}_d} \right\} _{d=1}^{m}=1,\left\{ U_{t}^{\mathrm{DG}_d} \right\} _{d=m+1}^{D}=0\right) .
		\end{equation}
 
    With the above strategy, we can first map each $m$ to $k_{m}$, and then choose the optimal $m^{*}$ as
     \begin{equation}
    	\label{m*}
    	m^* =\underset{m\in \hat{\mathcal{A}}^{\mathrm{SW}}}{\mathrm{arg}\max}Q\left( H_t,k_m,{a_{k_m}^{\mathrm{ED}}}^* \right) .
    \end{equation}
    
    Specifically, the HAFH-RDPG algorithm given in Algorithm \ref{alg-HAFH-DDPG}, \ref{alg-HAFH-DDPG-funcaction}, and \ref{alg-HAFH-DDPG-train} can be used to train the actor and critic networks by replacing the switching action $k$ with the simplified switching action $m$. However, in the eighth line of Algorithms \ref{alg-HAFH-DDPG}, $m_t^{(e)}$ should first be mapped to $k_t^{(e)}$ by the DG selection strategy before its execution to derive the reward and next state. Since each $m$ corresponds to a unique $k_{m}$, the reward $r(S_t,m,{a_{m}^{\mathrm{ED}}}^*)$ and transition probability $\mathrm{Pr}\left( H_{t+1}|H_t,m,{a_{m}^{\mathrm{ED}}}^* \right)$ can be derived by replacing $m$ with $k_{m}$ in the expressions. Finally, the optimal simplified switching action $m^{*}$ is mapped to the switching action $k_{m^{*}}$. 


    We now provide Theorem \ref{thm1} which elaborates that the two switching actions, i.e., $k^*$ derived by \eqref{k*} and $k_{m^{*}}$ with $m^*$ derived by \eqref{m*}, achieve the same performance. 
	

	\begin{thm}
		\label{thm1}
        Suppose that all the DG parameters are identical, the POMDPs with action space $\hat{\mathcal{A}}^{\mathrm{SW}}$ and $\mathcal{A}^{\mathrm{SW}}$ have the same optimal performance given that the above DG selection strategy is used to determine the $m$ DGs that are ON, i.e.,
        \begin{equation}
        	\label{thmeq}
	\underset{m\in \hat{\mathcal{A}}^{\mathrm{SW}}}{\max}Q\left( H_t,k_m,{a_{k_m}^{\mathrm{ED}}}^* \right) =\underset{k\in \mathcal{A}^{\mathrm{SW}}}{\max}Q\left( H_t,k,{a_{k}^{\mathrm{ED}}}^* \right) .
        \end{equation}
        
	\end{thm}

    \section{Numerical Analysis}
    In order to evaluate the effectiveness of the proposed algorithm, we perform experiments based on real-world data in an IoT-driven MG and discuss about the simulation results. All the experiments are performed on a Linux server, where the DRL algorithms are implemented in Tensorflow 1.14 using Python.

\subsection{Experimental Setup}
The MG system simulated in our experiments comprises three DGs, a PV panel, and a battery. The setting of three DGs is commonly adopted for experiments on isolated MGs \cite{sachs2016two,Li2019,Li2022a}. The parameters of the system model are listed in Table \ref{parameters}. Note that different values of coefficients $C_{\mathrm{R}}$, $C_{\mathrm{S}}$, and $C_{\mathrm{SR}}$ can result in different optimal policies, since varying these coefficients will change the weights of the DG power generation cost $c_{t}^{\mathrm{DG}_{d}}$, the start-up cost $c_{t}^{\mathrm{S\_DG}_{d}}$, running cost $c_{t}^{\mathrm{R\_DG}_{d}}$, and the spinning reserve cost $c_{t}^{\mathrm{SR\_DG}_{d}}$ in the operating cost $c_{t}^{\mathrm{OP}}$ according to \eqref{costDG}-\eqref{opcost}. Based on our setting, the DG generation cost occupies the largest proportion in the operating cost, followed by the spinning reserve cost, the running cost, and the start-up cost. Meanwhile, all the above costs are in the same order of magnitude. 

\subsubsection{Experimental data}
\begin{figure}[t]
	\centering
 \includegraphics[width=0.48\textwidth]{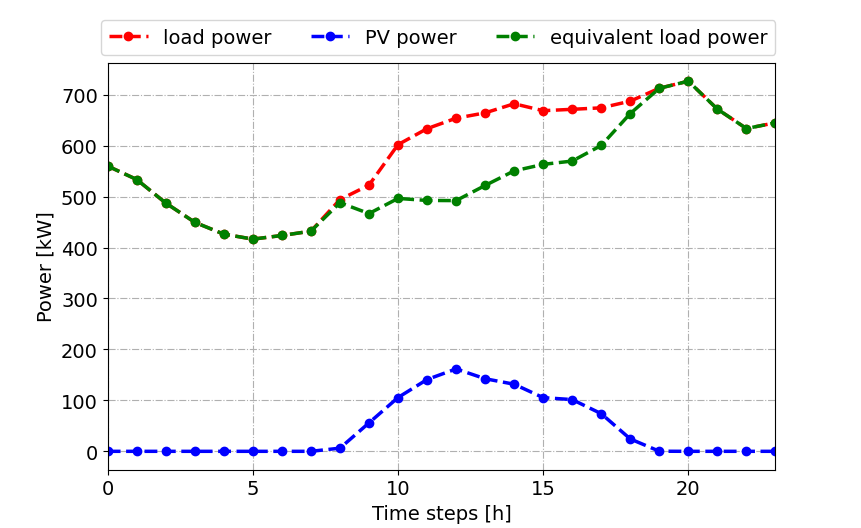}
	\caption{The trajectories of the PV power, load power, and equivalent load power of one typical day in the experimental data. }
	\label{dataset} 
\end{figure}

We use PV and load statistics in a real IoT-driven isolated MG as the experimental data. These data are collected by the sensors per hour. Therefore, the duration of a time step is set to one hour, i.e., $\Delta t=1h$. Each data at a time step include two values, representing the average load power and average PV power within the corresponding hour. Fig. \ref{dataset} shows the trajectories of the PV power, load power, and equivalent load power of one typical day in the experimental data. By observing the fluctuations over time in the curve, we find that the PV power is non-negative between 8 a.m. and 18 p.m. with a peak value of $180$ kW at around 12 p.m.; while the load power drops to the lowest value of $400$ kW between 4 a.m. and 6.a.m. and reaches peak value of $700$ kW between 18 p.m. and 21 p.m.. We regard one-day data as an episode, where the total number of time steps is $T=24$. In order to evaluate the capability of the proposed algorithm in dealing with inter-day power fluctuation, two types of data sets are designed for the experiment.
\begin{enumerate}
\item \emph{Type A}: The data of one single day randomly sampled from data set are used as both the training and test set.
\item \emph{Type B}: The data of $N$ continuous days randomly sampled from data set are used as the training set, while the test set comprises data of the next ($(N+1)$-th) day. 
\end{enumerate}

Type A can provide a benchmark scenario, where there is no uncertainty due to inter-day power fluctuation. Meanwhile, Type B is used to evaluate the performance of the proposed algorithm in a practical scenario. By comparing the performance of the proposed algorithm under the two data sets, we can evaluate whether the algorithm is able to learn a good policy from history data, which is generalized enough to work well for the next day with unknown PV generation and load demand. 

In addition, for both Type A and B, we randomly select five sets of data on different days and train the DRL algorithms in five runs. By comparing the performance of the proposed algorithm under different runs, the stability of the proposed algorithm can be evaluated. \par 

\subsubsection{DRL environment} 
In order to evaluate the capability of the proposed algorithm in dealing with inter-hour power fluctuation, two types of DRL environments are implemented in the experiments.
\begin{enumerate}
\item \emph{MDP}: The state information including PV and load data of the current time step is available at the beginning of a time step for making decisions.
\item \emph{POMDP}: Only the PV and load data of the previous $\tau$ time steps are available at the beginning of a time step, which are regarded as the history data for making decisions.
\end{enumerate}     
The MDP environment provides a benchmark scenario, where there is no uncertainty due to inter-hour power fluctuation. Meanwhile, the POMDP environment captures inter-hour power fluctuation in a practical scenario. Comparing the performance of the proposed algorithm under the two environments can prove whether the proposed algorithm is able to deal with partial observability problem and works well when PV and load data for the coming hour is unknown.\par
\subsubsection{Benchmark algorithms}
Several benchmark algorithms are selected for performance comparison with the proposed algorithm.
\begin{enumerate}
\item \emph{Myopic algorithm}: The action $A_t$ is directly optimized by minimizing the reward function $r_t(S_t, A_t)$ in \eqref{rt} without foreseeing the impact on future rewards at each time step $t\in \left\{ 1,2,...,T \right\} $. 
\item \emph{Discrete dynamic programming (DDP)} \cite{sachs2016two}: The finite-horizon value iteration algorithm in dynamic programming, where the states and actions are discretized. DDP guarantees to find the global optimum for the MDP after discretization, although performance loss could be incurred for the original MDP depending on the discretization granularity. 
\item \emph{Hybrid action finite-horizon DDPG (HAFH-DDPG)}: The version of HAFH-RDPG for MDP environment, where the LSTM layer is replaced by a fully-connected layer in actor and critic networks. 
\item \emph{DRQN} \cite{Hausknecht2015}: A popular value-based DRL algorithm for solving POMDP problem. To deploy DRQN, the action space needs to be discretized at proper granularity.
\end{enumerate}     

The hyper-parameters of the DRL algorithms are listed in Table \ref{DRLparameters}. \par

\begin{table}[t!]
	\centering
	\caption{Parameter configuration in the MG system model.}
	\begin{tabular}[b]{p{2.0cm}<{\raggedright}p{1.3cm}<{\raggedright}p{4.5cm}<{\raggedright}}
		\hline
		\textbf{Notations}&\textbf{Values}&\textbf{Description}\\
		\hline
		\specialrule{0em}{1pt}{1pt}
		$D$        & 3 &The number of DGs\\
		\specialrule{0em}{1pt}{1pt}
		$P_{\min}^{\mathrm{DG}}$ / $P_{\max}^{\mathrm{DG}}$ & 60kW / 300kW &The minimum/maximum power constraint of a DG\\
		\specialrule{0em}{1pt}{1pt}
		$P_{\mathrm{max}}^{\mathrm{E}}$  & 200kW &The power constraint of the battery\\
		\specialrule{0em}{1pt}{1pt}
		$E_{\mathrm{min}}$ / $E_{\mathrm{max}}$    & 24kWh / 600kWh &The minimum / maximum storage constraint of the battery\\
		\specialrule{0em}{1pt}{1pt}
		$\eta_{\mathrm{ch}}$/$\eta_{\mathrm{dis}}$  & 0.98 & The charging/discharging efficiency\\
		\specialrule{0em}{1pt}{1pt}
		$a_d$/$b_d$/$c_d$      & 0.0000381 / 0.1887 / 3.8571 &The quadratic / monomial / constant coefficient of cost curve\\
		\specialrule{0em}{1pt}{1pt}
		$C_{\mathrm{R}}$ / $C_{\mathrm{S}}$ / $C_{\mathrm{SR}}$  & 20 / 10 / 0.25  &The fixed running / start-up / spinning reserve cost coefficient\\
		\specialrule{0.05em}{2pt}{0pt}
	\end{tabular}
	\label{parameters}
\end{table}

\begin{table}[t!]
\centering
\caption{Hyper-parameters of the DRL algorithms.}
\begin{tabular}[b]{p{2.6cm}<{\centering}p{1.8cm}<{\centering}p{1.8cm}<{\centering}p{1.0cm}<{\centering}}
	\hline
	\textbf{Parameters} & \multicolumn{3}{c}{\textbf{Values}}\\
	\specialrule{0.05em}{3pt}{3pt}
	&\textbf{HAFH-DDPG} & \textbf{HAFH-RDPG} & \textbf{DRQN}  \\
	\specialrule{0em}{1pt}{1pt}
	\hline
	\specialrule{0em}{1pt}{1pt}
	Actor size       & 256,300,100 &128,128,64 & \textbackslash{} \\
	\specialrule{0em}{1pt}{1pt}
	Critic(Q) size       & 400,300,100 &128,128,64 & 256,300,100\\
	\specialrule{0em}{1pt}{1pt}
	\hline
	\specialrule{0em}{1pt}{1pt}
	Actor learning rate      & 5e-6 & 5e-6 & \textbackslash{}\\
	\specialrule{0em}{1pt}{1pt}
	Critic(Q) learning rate      & 5e-6 & 5e-6 & 1e-4\\		\specialrule{0em}{1pt}{1pt}
	\hline
	\specialrule{0em}{1pt}{1pt}
	History window & \textbackslash{} & 4 & 4\\		
	\specialrule{0em}{1pt}{1pt}
	Batch size & \multicolumn{2}{c}{128} & 64\\		
	\specialrule{0em}{1pt}{1pt}
	Training episodes & \multicolumn{3}{c}{15000}\\		
	\specialrule{0em}{1pt}{1pt}
	Reward scale  & \multicolumn{3}{c}{2e-3} \\
	\specialrule{0em}{1pt}{1pt}
	\hline
\end{tabular}
\label{DRLparameters}
\end{table}

\subsection{Experiment Results}
\subsubsection{Performance for test set}
Table \ref{result} summarizes the individual, average, maximum performance, as well as the standard error on five runs for all the algorithms. In each run, we executed $100$ complete episodes of $24$ time steps for each algorithm based on the test set, and obtain the individual performance by averaging the cumulative rewards over the $100$ episodes, where the cumulative reward of one episode is given by \eqref{cumulative reward}. Therefore, the performance corresponds to the average cumulative reward over $100$ test episodes, which reflects the optimization objective defined in \eqref{objective1}. The myopic, DDP, and HAFH-DDPG algorithms are implemented in MDP environment, while the HAFH-RDPG and DRQN algorithms are implemented in POMDP environment. Moreover, dataset Type A is used for all the benchmark algorithms, while both dataset Type A and B are used for HAFH-RDPG algorithms. \par


\begin{table*}[t]
	\centering
	\caption{The individual, average, maximum performance, as well as the standard error of the proposed and benchmark algorithms across five runs. In each run, we executed $100$ episodes of $24$ time steps. The individual performance is the average cumulative reward over $100$ test episodes.}
	\label{result}
	\begin{tabular}{|c|c|c|ccccccll|}
		\hline
		\multirow{2}{*}{\textbf{Environments}} & \multirow{2}{*}{\textbf{Algorithms}} & \multirow{2}{*}{\textbf{Dataset}} & \multicolumn{8}{c|}{\textbf{Performance}}                                                                                                                                                                                                                                                    \\ \cline{4-11} 
		&                                      &                                   & \multicolumn{1}{c|}{\textbf{Run 1}} & \multicolumn{1}{c|}{\textbf{Run 2}} & \multicolumn{1}{c|}{\textbf{Run 3}} & \multicolumn{1}{c|}{\textbf{Run 4}} & \multicolumn{1}{c|}{\textbf{Run 5}} & \multicolumn{1}{l|}{\textbf{Max}} & \multicolumn{1}{l|}{\textbf{Average}} & \textbf{Std Error} \\ \hline
		\multirow{3}{*}{MDP}                   & Myopic                               & \multirow{5}{*}{Type A}           & \multicolumn{1}{c|}{-12.8175}       & \multicolumn{1}{c|}{-13.5276}       & \multicolumn{1}{c|}{-14.9506}       & \multicolumn{1}{c|}{-16.8130}       & \multicolumn{1}{c|}{-13.2446}       & \multicolumn{1}{l|}{-12.8175}     & \multicolumn{1}{l|}{-14.2707}         & 1.6311             \\ \cline{2-2} \cline{4-11} 
		& DDP                                  &                                   & \multicolumn{1}{c|}{-11.2421}       & \multicolumn{1}{c|}{-12.1325}       & \multicolumn{1}{c|}{-12.5870}       & \multicolumn{1}{c|}{-15.1273}       & \multicolumn{1}{c|}{-11.7430}       & \multicolumn{1}{c|}{-11.2421}     & \multicolumn{1}{l|}{-12.5664}         & 1.5148             \\ \cline{2-2} \cline{4-11} 
		& HAFH-DDPG                            &                                   & \multicolumn{1}{c|}{-11.1819}       & \multicolumn{1}{c|}{-11.8509}       & \multicolumn{1}{c|}{-12.4679}       & \multicolumn{1}{c|}{-14.9620}       & \multicolumn{1}{c|}{-11.5876}       & \multicolumn{1}{c|}{-11.1819}     & \multicolumn{1}{l|}{-12.4101}         & 1.5011             \\ \cline{1-2} \cline{4-11} 
		\multirow{5}{*}{POMDP}                 & DRQN                                 &                                   & \multicolumn{1}{c|}{-12.0895}       & \multicolumn{1}{c|}{-12.9763}       & \multicolumn{1}{c|}{-14.1435}       & \multicolumn{1}{c|}{-15.8547}       & \multicolumn{1}{c|}{-12.4533}       & \multicolumn{1}{c|}{-12.0895}     & \multicolumn{1}{l|}{-13.5035}         & 1.5262             \\ \cline{2-2} \cline{4-11} 
		& \multirow{4}{*}{HAFH-RDPG}           &                                   & \multicolumn{1}{c|}{-11.2208}       & \multicolumn{1}{c|}{-12.2243}       & \multicolumn{1}{c|}{-12.7963}       & \multicolumn{1}{c|}{-15.1054}       & \multicolumn{1}{c|}{-11.6984}       & \multicolumn{1}{c|}{-11.2208}     & \multicolumn{1}{l|}{-12.6090}         & 1.5142             \\ \cline{3-11} 
		&                                      & Type B, $N=7$                     & \multicolumn{1}{c|}{-11.4238}       & \multicolumn{1}{c|}{-12.4536}       & \multicolumn{1}{c|}{-13.0564}       & \multicolumn{1}{c|}{-15.3693}       & \multicolumn{1}{c|}{-11.9409}       & \multicolumn{1}{c|}{-11.4238}     & \multicolumn{1}{l|}{-12.8488}         & 1.5336             \\ \cline{3-11} 
		&                                      & Type B, $N=14$                    & \multicolumn{1}{c|}{-11.4021}       & \multicolumn{1}{c|}{-12.3575}       & \multicolumn{1}{c|}{-13.3536}       & \multicolumn{1}{c|}{-15.2703}       & \multicolumn{1}{c|}{-12.0685}       & \multicolumn{1}{c|}{-11.4021}     & \multicolumn{1}{l|}{-12.8904}         & 1.5044             \\ \cline{3-11} 
		&                                      & Type B, $N=21$                    & \multicolumn{1}{c|}{-11.6264}       & \multicolumn{1}{c|}{-12.5459}       & \multicolumn{1}{c|}{-13.5580}       & \multicolumn{1}{c|}{-15.5341}       & \multicolumn{1}{c|}{-12.3246}       & \multicolumn{1}{c|}{-11.6264}     & \multicolumn{1}{l|}{-13.1178}         & 1.5176             \\ \hline
	\end{tabular}
\end{table*}

\subsubsection{Convergence Properties}
\begin{figure}[t]
\centering
	\includegraphics[height=5.6cm,width=0.48\textwidth]{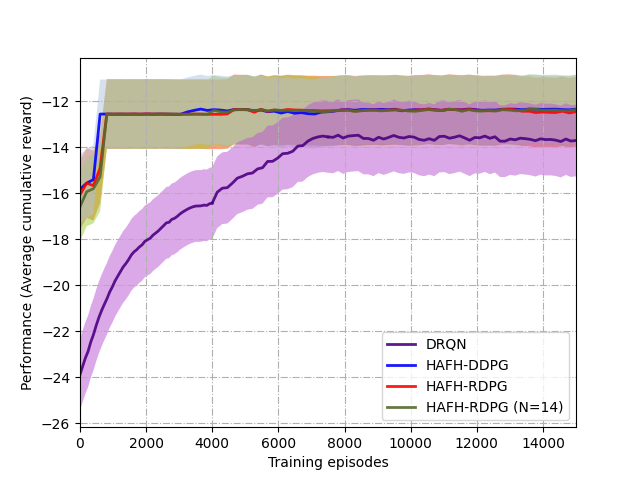}
\caption{The average performance curves of three DRL algorithms across five runs in the training process. The ordinate corresponds to the performance, i.e., the average cumulative reward over $10$ test episodes. The shaded areas indicate the standard errors of the three algorithms.}
\label{training-result} 
\end{figure}

Fig. \ref{training-result} shows the average performance curves of the DRL algorithms across five runs, which is obtained by evaluating the policies periodically during training without noise. In specific, during the training process, we ran $10$ test episodes based on the current model parameters every $100$ training episodes in order to accurately represent the training results. The abscissa is the number of training episodes and the ordinate is the performance of DRL algorithms, which denotes the average cumulative reward  over $10$ test episodes. The shaded areas indicate the standard errors of the algorithms across the five runs. Moreover, for HAFH-DDPG and HAFH-RDPG algorithms, we only show the performance curves of the first time step $t=1$, as the performance curves for all $T$ time steps are similar.

\subsubsection{Energy Dispatch Results}
\begin{figure*}[ht!]
\centering
\subfigure[energy scheduling of myopic]{
	\label{powermyopic} 
	\includegraphics[height=4.2cm,width=0.49\textwidth]{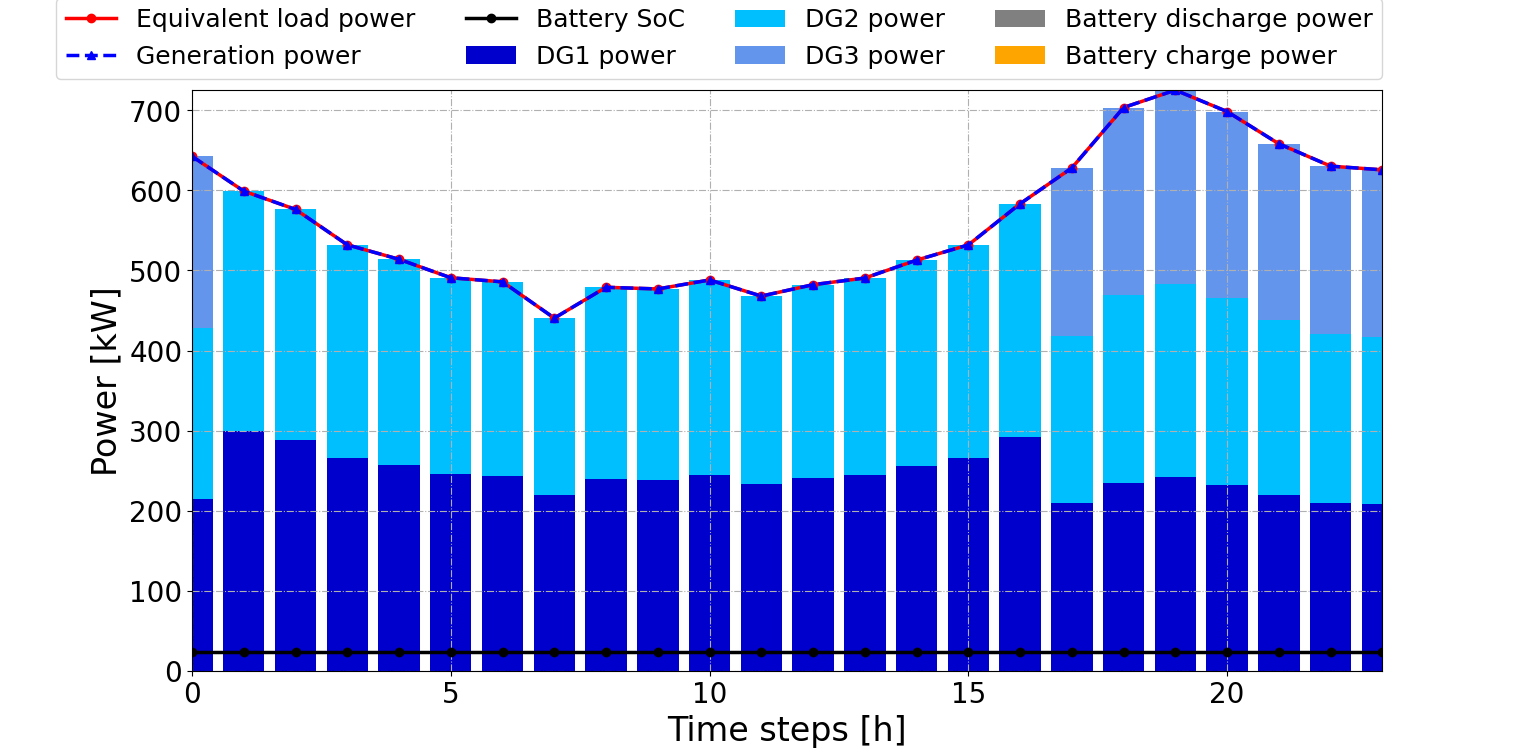}}
\subfigure[costs of myopic]{
	\label{costmyopic} 
	\includegraphics[height=4.2cm,width=0.49\textwidth]{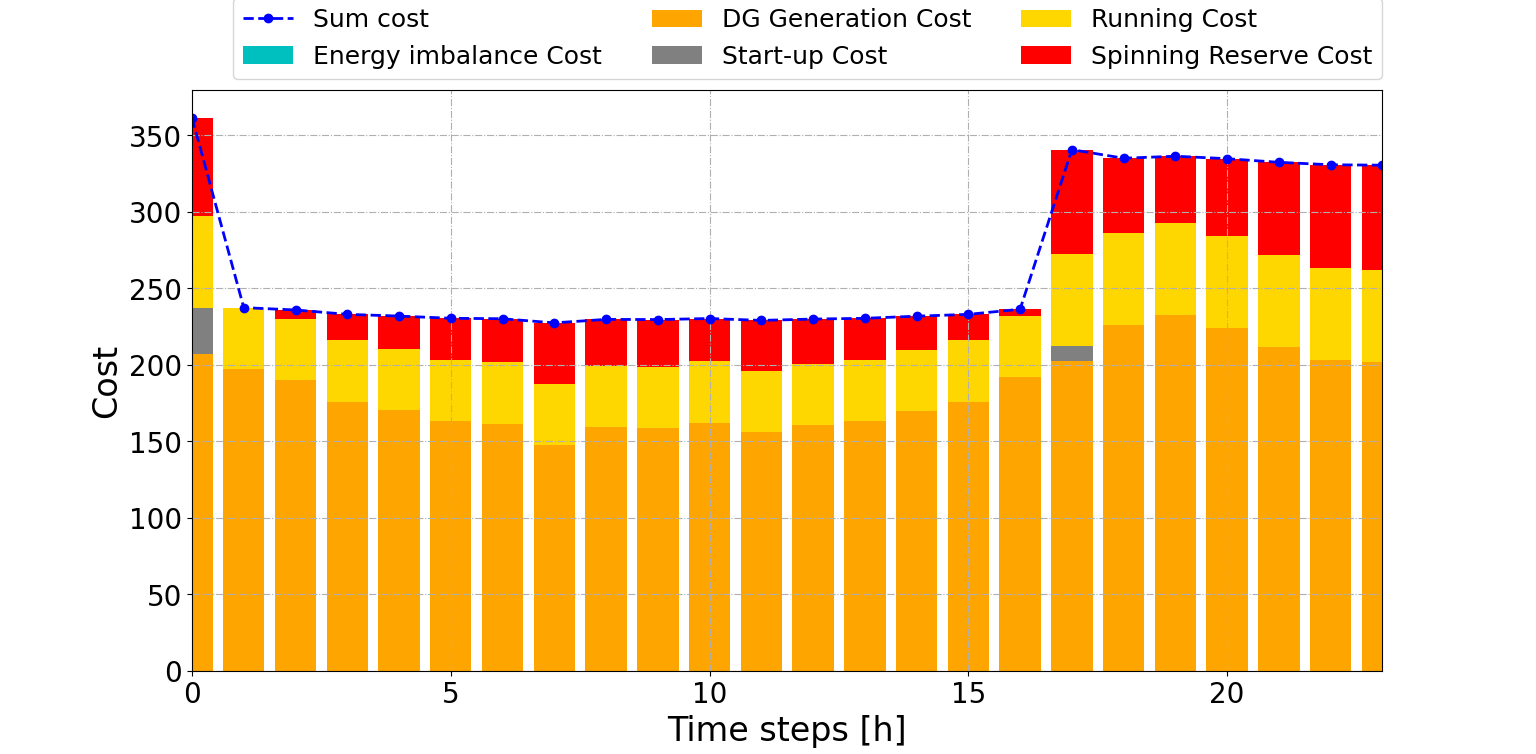}}
\subfigure[energy scheduling of DDP]{
	\label{powerDDP} 
	\includegraphics[height=4.2cm,width=0.49\textwidth]{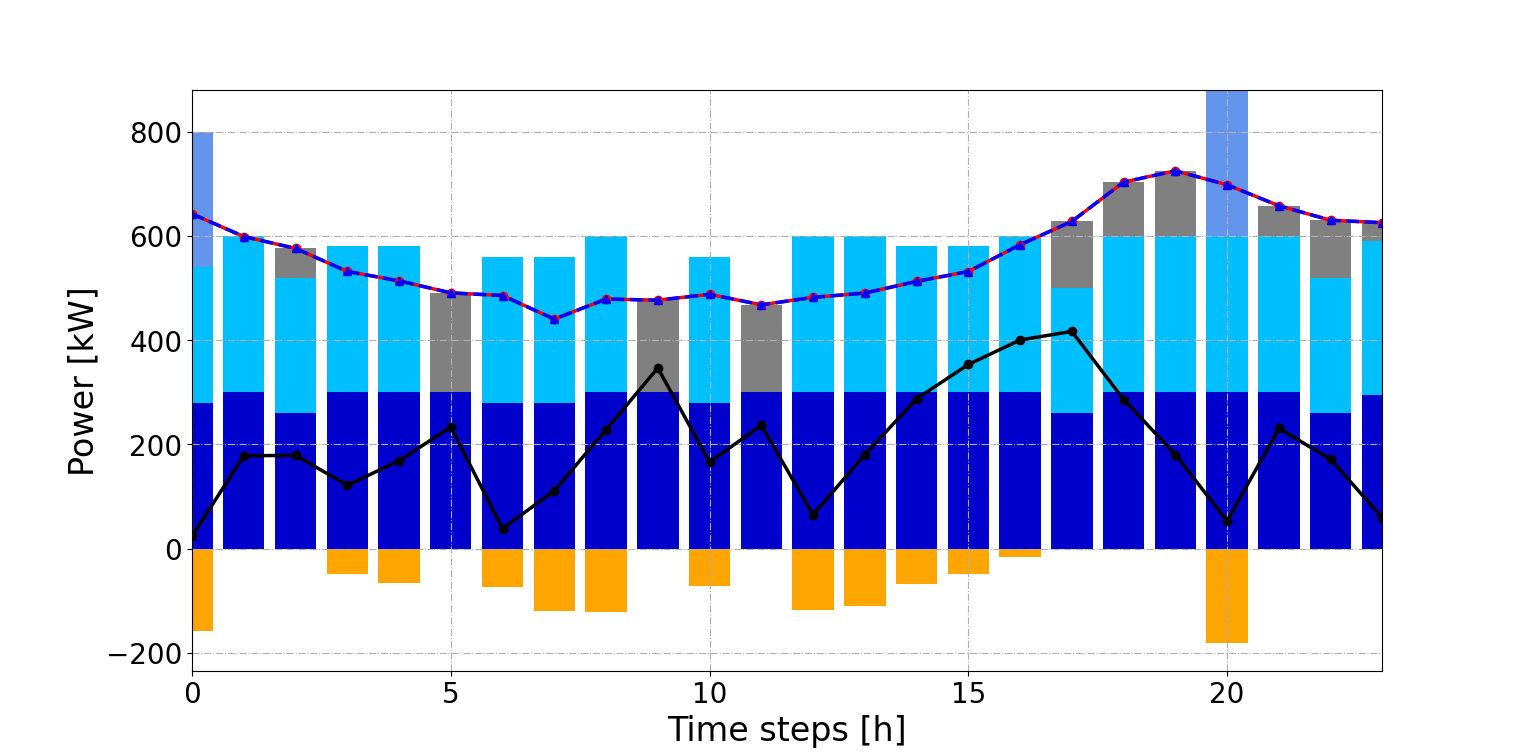}}
\subfigure[costs of DDP]{
	\label{costDDP} 
	\includegraphics[height=4.2cm,width=0.49\textwidth]{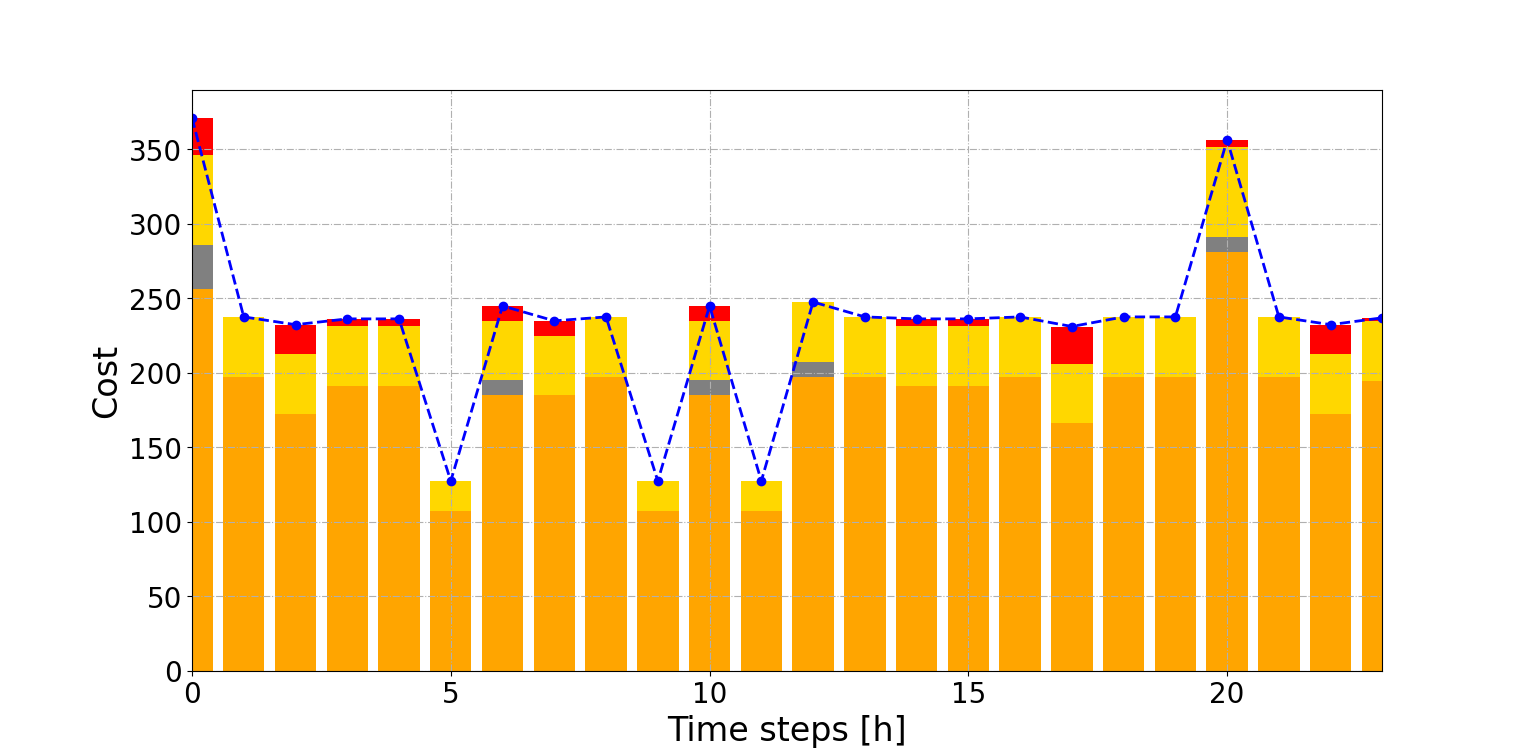}}
\subfigure[energy scheduling of HAFH-DDPG]{
	\label{powerDDPG} 
	\includegraphics[height=4.2cm,width=0.49\textwidth]{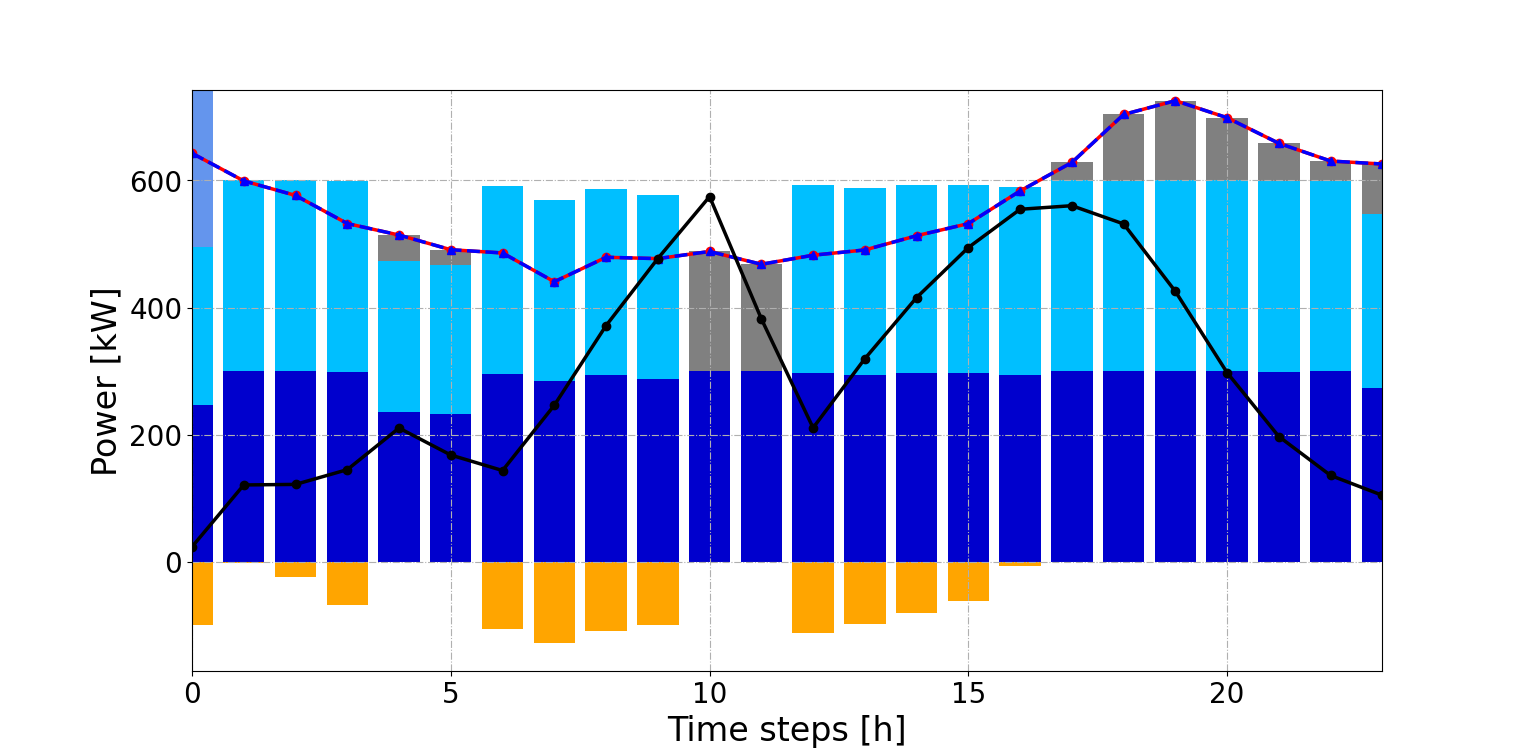}}
\subfigure[costs of HAFH-DDPG]{
	\label{costDDPG} 
	\includegraphics[height=4.2cm,width=0.49\textwidth]{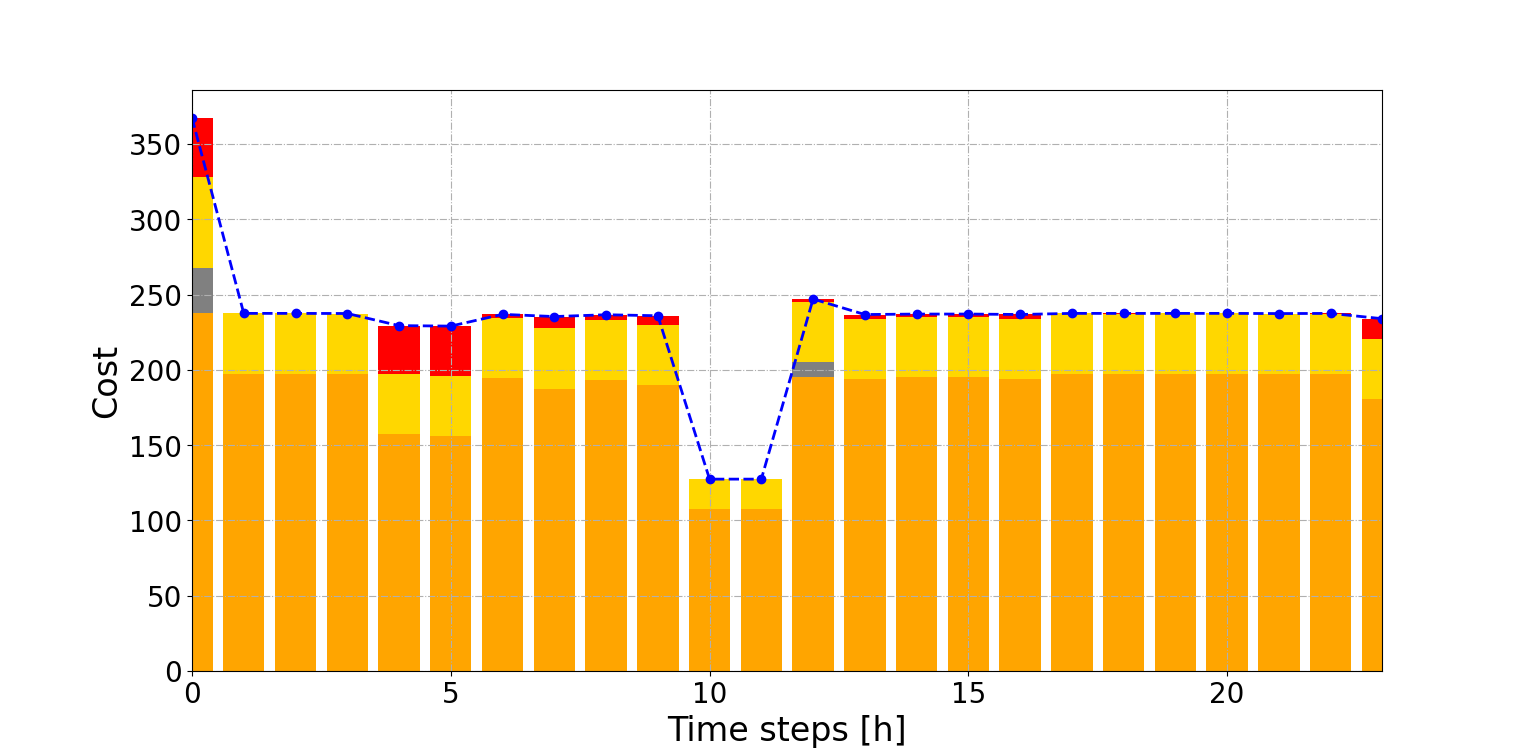}}
\subfigure[energy scheduling of HAFH-RDPG]{
	\label{powerRDPG} 
	\includegraphics[height=4.2cm,width=0.49\textwidth]{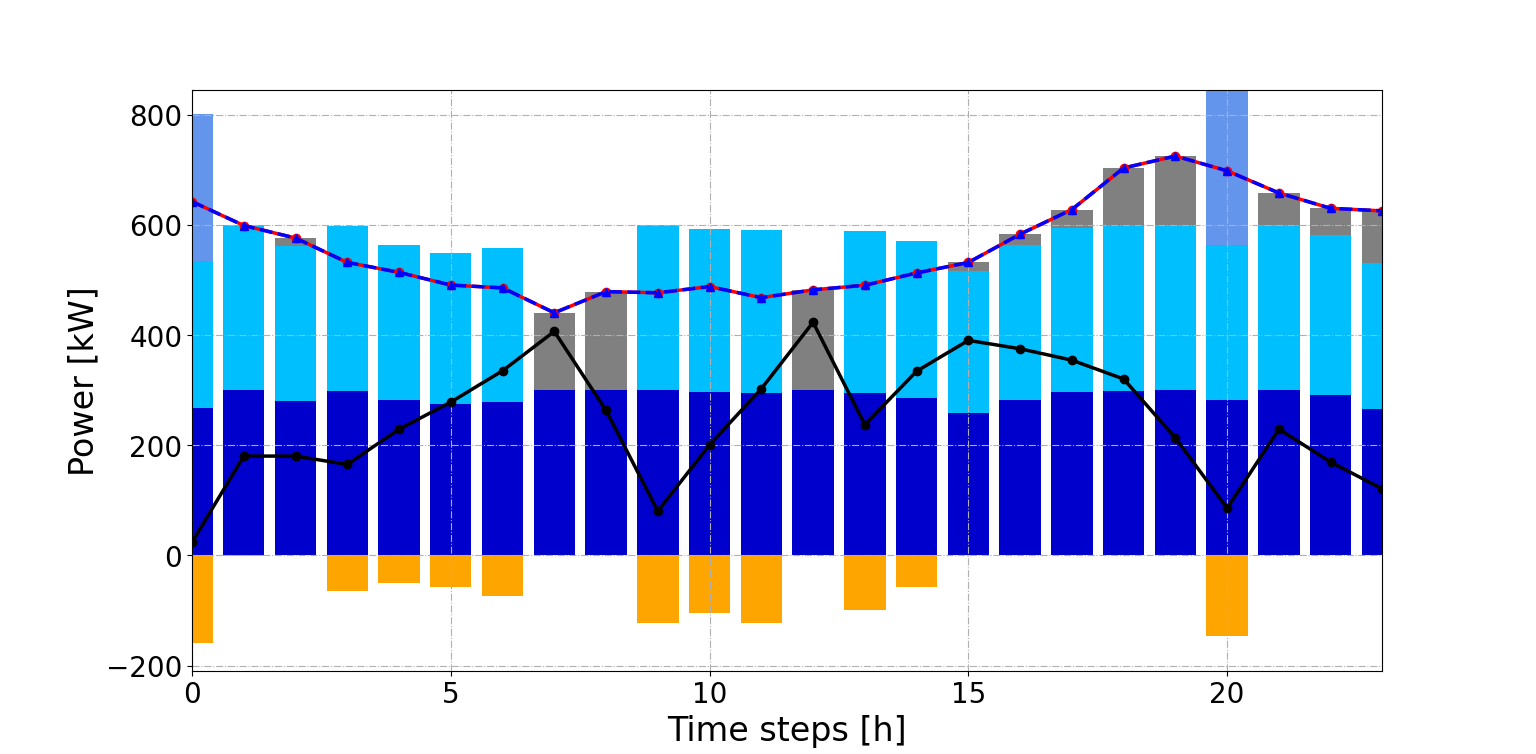}}
\subfigure[costs of HAFH-RDPG]{
	\label{costRDPG} 
	\includegraphics[height=4.2cm,width=0.49\textwidth]{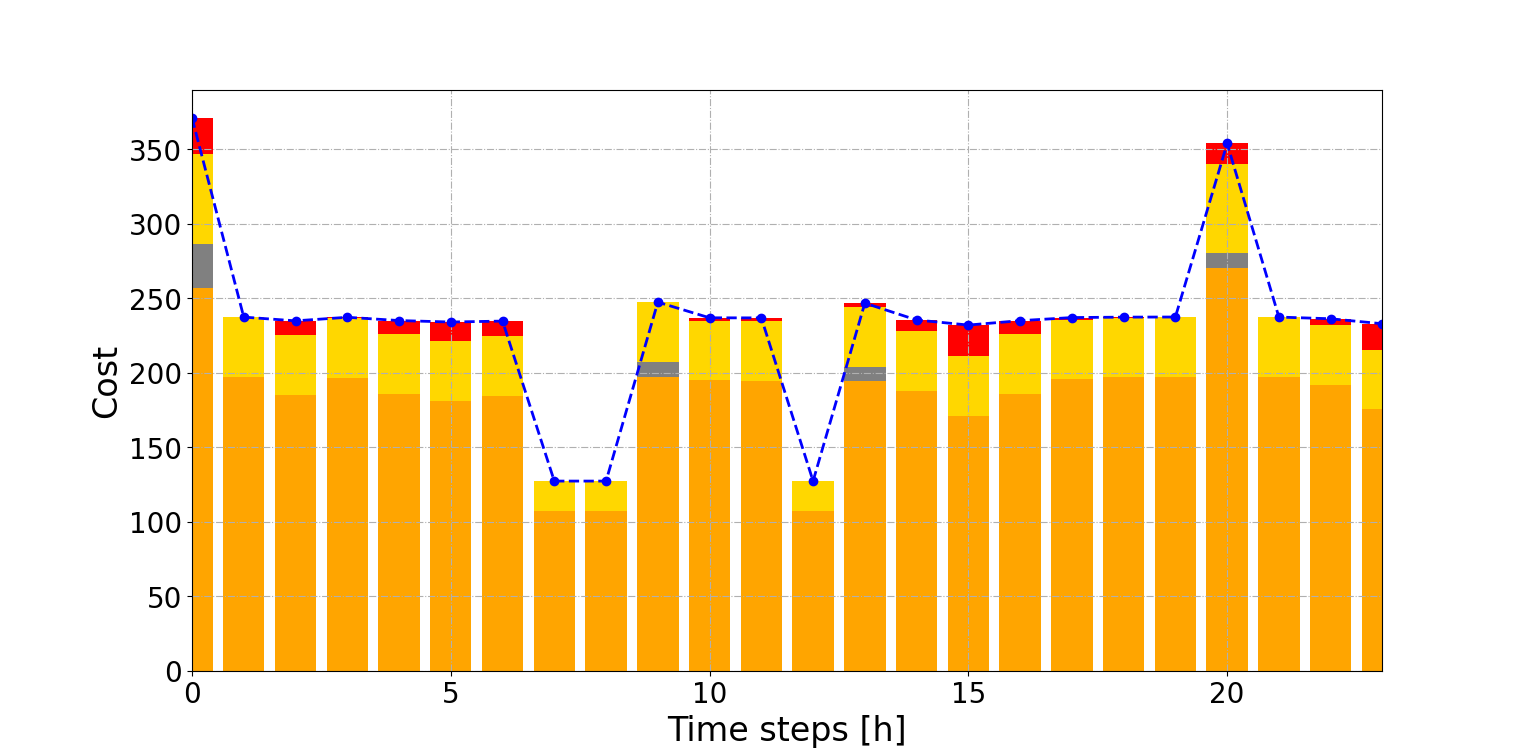}}


\end{figure*}

\begin{figure}[t]
	\centering
		\includegraphics[height=4.5cm,width=0.52\textwidth]{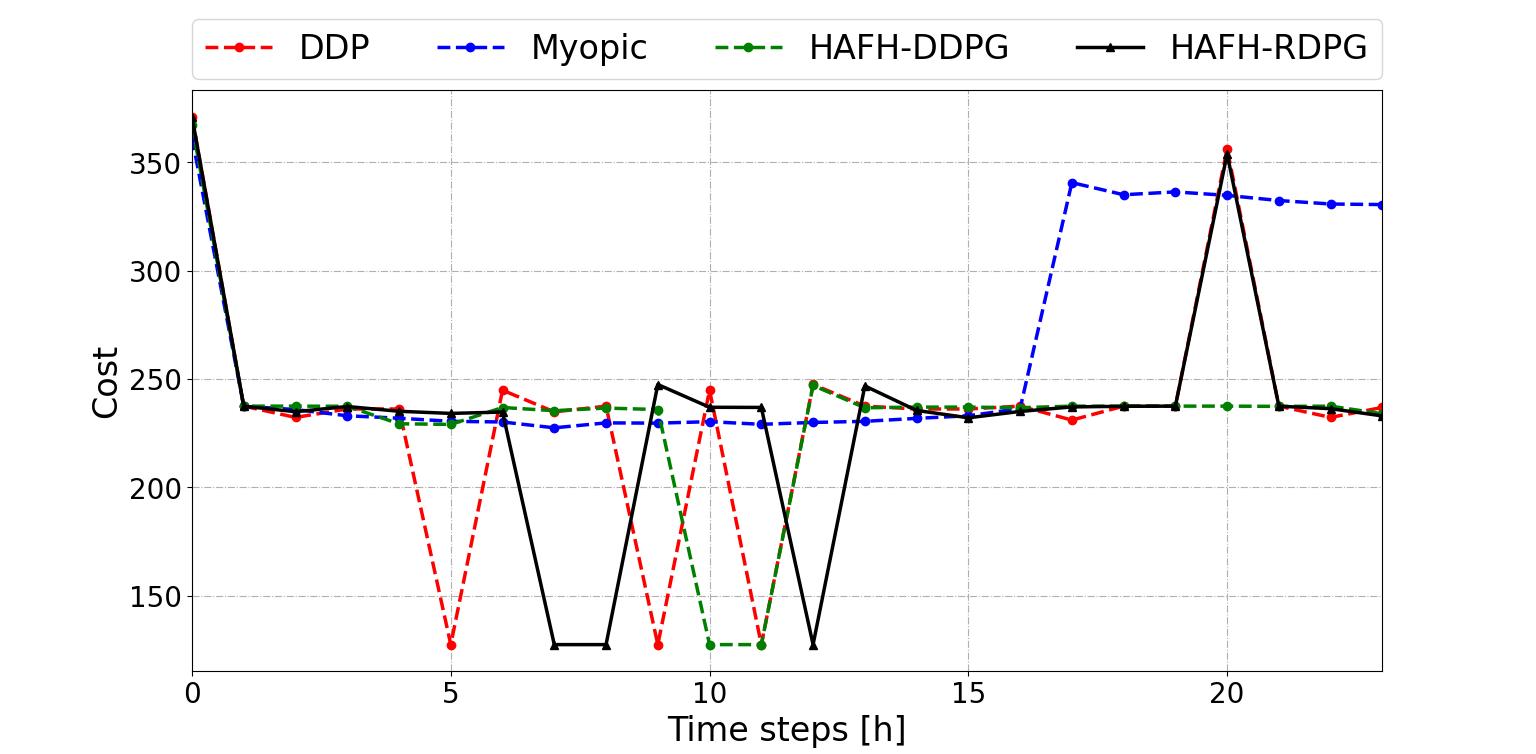}
	\caption{The sum cost trajectories for a specific test episode to compare the performance of different algorithms. }
	\label{sumcost} 
\end{figure}

 Fig. \ref{trajectory} illustrates the energy scheduling decisions and corresponding costs of different algorithms over one day based on the test set of run 1. Due to space limitation, only the results for myopic, DDP, HAFH-DDPG, and HAFH-RDPG (Type A) are shown. In Fig. \ref{powermyopic}, \ref{powerDDP}, \ref{powerDDPG}, \ref{powerRDPG}, three curves are displayed corresponding to the equivalent load trajectory, generation power trajectory, and battery trajectory, respectively, in each figure. The generation power value at time step $t$ is derived by $\sum_{d=1}^{D}P_{t}^{\mathrm{DG}_{d}}- P_{t}^{\mathrm{E}}$, i.e., the total power of DGs after charging/discharging of battery. Moreover, the charge or discharge power of battery and the power generated by each of the three DGs are also shown through the bar chart. In Fig. \ref{costmyopic}, \ref{costDDP}, \ref{costDDPG}, \ref{costRDPG}, the sum cost trajectory is displayed in each figure, where the sum cost at time step $t$ is the negative reward $-r(S_t,A_t)$. Moreover, we use the bar chart to visualize the individual values of different costs in detail. Finally, Fig.\eqref{sumcost} displays the sum cost trajectories of the different algorithms in the same figure to clearly show the sum cost comparisons.



\begin{figure*}[ht!]
\caption{The equivalent load, generation power, battery and sum cost trajectories of various algorithms for a specific test episode on run 1. The charge or discharge power of the battery, the power generated by each of the three DGs, and the individual costs are shown through the bar chart.}
\label{trajectory} 
\end{figure*}

\subsection{Analysis and Discussion} 
\subsubsection{Capability in handling uncertainty due to inter-hour power fluctuation}
We focus on the performance of myopic, DDP, HAFH-DDPG, and HAFH-RDPG algorithms when dataset Type A is used. Note that the benchmark algorithms are all implemented in MDP environment, where the PV and load data of the current time step is assumed to be available, while the proposed HAFH-RDPG algorithm is implemented in POMDP environment, where the PV and load data of the past four time steps are used to determine the actions. It can be observed in Table \ref{result} that HAFH-DDPG has the largest maximum performance, followed by HAFH-RDPG, DDP and myopic algorithm. Moreover, HAFH-DDPG also achieves the best average performance across five runs among all the algorithms, while DDP and HAFH-RDPG (for Type A) have the second and third best average performance. The myopic algorithm has the worst average performance on five runs. Finally, HAFH-DDPG achieves the lowest standard error, followed by HAFH-RDPG, DDP and myopic algorithm.

For each run, the individual performance of HAFH-DDPG is consistently better than those of the other algorithms. As HAFH-DDPG is the version of HAFH-RDPG in MDP environment, its high performance indicates that our proposed algorithm works well when there is no uncertainty due to inter-hour power fluctuation, and thus no performance degradation due to partial observation. Moreover, although HAFH-RDPG is applied in the POMDP environment, it outperforms DDP in the MDP environment on run 1, run 4 and run 5, which demonstrates that our proposed algorithm can exploit history data to deal with the uncertainty due to inter-hour power fluctuation and make efficient decisions. Although the performance of DDP is comparable to that of HAFH-RDPG, it is trained in the MDP environment ignoring the partial observable problem due to the unavailable PV generation and load demand data of the next hour. When prediction errors for the next-hour data occur in realistic scenarios, it can be expected that the performance of DDP will degrade and become worse than that of HAFH-RDPG.  More importantly, the computational complexity of DDP is much higher than that of HAFH-RDPG. 
\noindent Although theoretically, DDP can obtain the optimal solution when the discretization granularity of state and action space is infinitely small, it is not a practical option as the computation complexity $O(|\mathcal{S}|^2*|\mathcal{A}|)$ depends on the state and action space dimensions. In order to achieve the presented performance in our experiment, we train DDP for $100$ iterations and the optimal value function is updated by over $2.7$ billion times in each iteration. Overall, the running time of DDP is approximately $11$ times that of HAFH-RDPG. Finally, the standard errors across the five runs are mainly due to the differences in PV and load data on different days instead of the instability of DRL algorithms. This is corroborated by the fact that the standard error of the myopic algorithm is higher than those of the DRL algorithms. It can be observed that our proposed algorithm performs stably in both MDP and POMDP environment, as the HAFH-DDPG and HAFH-RDPG algorithms achieve the first and second lowest standard errors.

Focusing on the convergence properties in Fig. \ref{training-result}, it can be observed that HAFH-DDPG has the fastest convergence speed and converges at approximately $600$ episodes. Despite being in the POMDP environment, HAFH-RDPG also converges fast at approximately $800$ episodes. Moreover, the shaded areas of HAFH-DDPG and HAFH-RDPG are similar and have a large overlap, which indicates that the stability of the proposed algorithms is comparable in both MDP and POMDP environment.

For the ED results in Fig. \ref{trajectory}, we find that the curves of load and generation power match exactly for all the algorithms, which means the basic energy balance requirement can be met. A closer examination shows that the generation power and battery trajectories of DDP (Fig. \ref{powerDDP}), HAFH-DDPG (Fig. \ref{powerDDPG}), and HAFH-RDPG (Fig. \ref{powerRDPG}) are very similar. These algorithms adopt similar strategies which try to keep the DGs that are ON generating very high power, which can largely reduce spinning reserve cost. The surplus power is charged to the battery so that one or more DGs can be turned off to reduce running cost when the battery has enough power to meet the load demand. As a result, two DGs are ON at most time steps, and sometimes only one ON DG is needed during the morning and noon hours. In addition, it can be observed visually that only two DGs are used in HAFH-DDPG except for the first time step as shown in Fig. \ref{powerDDPG}, which leads to the lowest costs as shown in Fig. \ref{sumcost}. This conclusion shows that the MDP version of the proposed algorithm achieves the best results, which is also verified by Table \ref{result}. Most importantly, although the PV and load power of the current time step cannot be obtained by HAFH-RDPG, Fig. \ref{powerRDPG} shows that this algorithm still outperforms DDP in energy scheduling on run 1. Compared with DDP, HAFH-RDPG only incurs the start-up cost three times at 9 a.m., 13 p.m. and 20 p.m., which reduces the loss of frequently changing switching states. Finally, compared with the above policies, the myopic algorithm is unable to take advantage of the battery to charge in advance for the evening peak and for energy shifting. Therefore, it has to turn on three DGs after 17 p.m., which leads to great spinning reserve cost and DG generation cost as shown in Fig. \ref{sumcost}.

\subsubsection{Capability in handling uncertainty due to inter-day power fluctuation} 
We compare the performance of HAFH-RDPG when both dataset Type A and B are used, respectively. For Type B, we set the number of past days for training, i.e., $N$, to be $7$, $14$, and $21$, respectively, to prevent the agent from over-fitting to the statistics of one particular day. Table \ref{result} shows that the best average performance for Type B ($N=7$) is only $1.90\%$ lower than that for Type A and the best maximum performance for Type B ($N=14$) is only $1.62\%$ lower than that for Type A, demonstrating that the proposed algorithm has the capability in handling uncertainty due to inter-day power fluctuation. Comparing the performance for Type B when different numbers of past days $N$ are used for training, it can be observed that the individual performance for $N=21$ on each run is always the worst. The individual performance for $N=7$ is better than those for $N=14$ on run 1, run 2 and run 4, while the opposite is true on run 3 and run 5. Overall, the average performance for $N=21$ is $2.09\%$ and $1.76\%$ lower than that for $N=7$ and $N=14$, respectively. This is because when $N$ is smaller, the training data are more recent, and there is a higher probability that the statistics of PV and load data are more similar to those of the test data. Meanwhile, in the rare case when the test data is an outliner, whose statistical properties are much different from those of data in the recent past, increasing $N$ can make the training data more general and thus improve the performance of the trained policy on the test data. Therefore, the setting for $N$ should strike a good trade-off, and both $N=7$ and $N=14$ are appropriate in this case.
	
Fig. \ref{training-result} shows that the convergence speed and stability of HAFH-RDPG is very similar when trained using dataset Type A or B, which is as expected. In addition, notice that DDP requires predictive models to handle the uncertainty due to inter-day power fluctuation, while HAFH-RDPG directly learns a policy from the history data. The resultant robustness to prediction model error is also an advantage of the proposed algorithm over DDP.




\subsubsection{Capability in dealing with hybrid action space} 
We compare the performance of DRQN and HAFH-RDPG, where both algorithms work in POMDP environment. However, DRQN uses discretization to deal with the hybrid action space, while HAFH-RDPG uses the proposed method. Table \ref{result} shows that HAFH-RDPG always has better individual performance on each run than DRQN. The maximum performance of HAFH-RDPG is $7.74\%$ larger than that of DRQN, and the average performance of HAFH-RDPG is $7.09\%$ larger than that of DRQN, demonstrating the superior capability of HAFH-RDPG in dealing the hybrid action space. The standard error of DRQN is $0.79\%$ larger than that of HAFH-RDPG, which indicates that the proposed algorithm is more stable than DRQN.  \par  

Focusing on the convergence properties, Fig. \ref{training-result} shows that HAFH-RDPG converges much faster and with smaller fluctuation than DRQN. The former converges at approximately $800$ training episodes, while the latter converges at approximately $7,500$ episodes. This is because HAFH-RDPG iteratively train to solve the one-period POMDP problem for each time step using RDPG with fixed target, which makes our proposed algorithm much easier to converge and more stable. 

\section{Conclusion}
In this paper, we have studied the energy scheduling issue with low operation cost in IoT-driven isolated smart MG systems by fully exploiting the renewable energy. The DRL approach has been adopted to handle the uncertainty of PV generation and load demand. A finite-horizon POMDP model has been developed considering the spinning reserve. The HAFH-RDPG algorithm has been proposed, which overcomes the challenge of learning optimal policy with discrete-continuous hybrid action space. Finally, experiments have been performed to demonstrate that the proposed algorithm can efficiently tackle with the uncertainty due to inter-hour and inter-day power fluctuation, and can achieve better performance than the other benchmark algorithms. In the future, this work will be further extended by optimizing energy dispatch of network PV panels jointly with the scheduling of DGs, which can help to minimize reverse power flow in the isolated MGs. Due to the distributed deployment of residential PV, multi-agent DRL and federated DRL can be leveraged. 




\begin{appendices}
\section{Proof for Theorem \ref{thm_optimal}}
	\begin{proof}
The optimal policy for the objective in \eqref{objective} achieves the optimal Q value, i.e.,

  \begin{equation}
 \label{sameswaction1}
   Q\left(H_t,k^*,{a_{k^*}^{\mathrm{ED}}}^*\right) =\underset{k\in \left\{ 1,...,2^D \right\}, a_{k}^{\mathrm{ED}}\in \mathcal{A} _{k}^{\mathrm{ED}}}{\max}Q\left( H_t,k,a_{k}^{\mathrm{ED}} \right) ,
 \end{equation}

 Combining \eqref{akED*} and \eqref{k*}, the policy derived by the two-step framework also achieves the optimal Q value, i.e.,

  \begin{align}
 \label{sameswaction2}
   Q\left(H_t,\hat{k}^*,{a_{\hat{k}^*}^{\mathrm{ED}}}^*\right) & =\underset{k\in \left\{ 1,...,2^D \right\}}{\max}\underset{a_{k}^{\mathrm{ED}}\in \mathcal{A} _{k}^{\mathrm{ED}}}{\max}Q\left( H_t,k,a_{k}^{\mathrm{ED}} \right) , \IEEEnonumber\\
   & = \underset{k\in \left\{ 1,...,2^D \right\}, a_{k}^{\mathrm{ED}}\in \mathcal{A} _{k}^{\mathrm{ED}}}{\max}Q\left( H_t,k,a_{k}^{\mathrm{ED}} \right) \IEEEnonumber\\
   &=Q\left(H_t,k^*,{a_{k^*}^{\mathrm{ED}}}^*\right)
 \end{align}
 
   \end{proof}

 


	\section{Proof for Theorem \ref{thm1}}
	\begin{proof}

		Our proof is divided into two steps. The first step is to prove that both $k^{*}$ and $k_{m^{*}}$ can achieve the maximum immediate reward at any time step $t$. According to the definitions in \eqref{costUS}, \eqref{costDG}, \eqref{costru} and \eqref{costSR}, it is obvious that DGs have translatable symmetry for the supply-demand balance cost, power generation cost, running cost, and spinning reserve cost. Therefore, the sum of these costs is the same for any $k\in {\mathcal{A}}_{m}^\mathrm{SW}$, i.e.,
		\begin{equation}
			\label{threecost}
			\begin{split}
				c_{t,k_1}^{\mathrm{US}}+\sum_{d=1}^D{(c_{t,k_1}^{\mathrm{DG}_d}+c_{t,k_1}^{\mathrm{R}\_\mathrm{DG}_d}+c_{t,k_1}^{\mathrm{SR}\_\mathrm{DG}_d})}\\
				=c_{t,k_2}^{\mathrm{US}}+\sum_{d=1}^D{(c_{t,k_2}^{\mathrm{DG}_d}+c_{t,k_2}^{\mathrm{R}\_\mathrm{DG}_d}+c_{t,k_2}^{\mathrm{SR}\_\mathrm{DG}_d})}, \\
				\forall {k_1,k_2}\in {\mathcal{A}}_{m}^\mathrm{SW}.
			\end{split}
		\end{equation}
		\noindent In addition, the DG selection strategy can ensure that the maximum number of ON DGs at time step $t$ are selected to be ON at time step $t+1$. In other words, the strategy minimizes the number of DGs whose $B_{t}^{\mathrm{DG}_{d}}=0$ and $U_{t}^{\mathrm{DG}_{d}}=1$. According to \eqref{eq2} and \eqref{costst}, the DG selection strategy in \eqref{atUCstr} leads to the switching action ${k}_m$ in ${\mathcal{A}}_{m}^\mathrm{SW}$ that minimizes the start-up cost, i.e.,
		\begin{equation}
			\label{stcost}
		{k}_m=\underset{k\in {\mathcal{A}}_{m}^\mathrm{SW}}{\mathrm{arg}\min}\left( \sum_{d=1}^D{c_{t,k}^{\mathrm{S}\_\mathrm{DG}_d}} \right) .
     	\end{equation}
		Therefore, according to the reward definition in \eqref{rt}, ${k}_m$ derived from the strategy maximizes the reward according to \eqref{threecost} and \eqref{stcost}, i.e.,
		\begin{equation}
			\label{kappa*}
			{k}_m=\underset{k\in {\mathcal{A}}_{m}^\mathrm{SW}}{\mathrm{arg}\max}\left( r(S_t,k,{a_{k}^{\mathrm{ED}}}^*) \right) .
		\end{equation}
		Thus, we can derive
		\begin{equation}
			\label{rtequal}
			\begin{split}
				&\underset{k\in \mathcal{A} ^{\mathrm{SW}}}{\max}\left( r(S_t,k,{a_{k}^{\mathrm{ED}}}^*) \right)\overset{\left( a \right)}{=}\underset{m\in \hat{\mathcal{A}}^{\mathrm{SW}}}{\max}\underset{k \in \mathcal{A} _{m}^{\mathrm{SW}}}{\max}\left( r(S_t,k,{a_{k}^{\mathrm{ED}}}^*) \right) \\
				&\overset{\left( b \right)}{=}\underset{m\in \hat{\mathcal{A}}^{\mathrm{SW}}}{\max}\left( r(S_t,k_m,{a_{k_m}^{\mathrm{ED}}}^*) \right), 
			\end{split}		 	
		\end{equation}
		\noindent where (a) is due to \eqref{cup}; and (b) follows from \eqref{kappa*}. 
		 
		In the second step, we prove \eqref{thmeq} using induction by the Bellman equation \eqref{bellman}. For the last time step $T$, we have 
		\begin{equation}
			\label{Bellmanlast}
			Q_T\left( H_T,k,{a_{k}^{\mathrm{ED}}}^* \right) =\underset{S_T|H_T}{\mathbb{E}}\left[r\left( S_T,k,{a_{k}^{\mathrm{ED}}}^* \right)\right] ,\forall k\in \mathcal{A} ^{\mathrm{SW}}.
		\end{equation}
		Thus, combining \eqref{rtequal} and \eqref{Bellmanlast}, we have
		\begin{equation}
			\underset{m\in \hat{\mathcal{A}}^{\mathrm{SW}}}{\max}Q_T\left( H_T,k_m,{a_{k_m}^{\mathrm{ED}}}^* \right) =\underset{k\in \mathcal{A}^{\mathrm{SW}}}{\max}Q_T\left( H_T,k,{a_{k}^{\mathrm{ED}}}^* \right) .
		\end{equation}
		Then, we assume that 
		\begin{equation}
			\label{thmeqt+1}
			\begin{split}
				\underset{m^{'}\in \hat{\mathcal{A}}^{\mathrm{SW}}}{\max}Q_{t+1}\left( H_{t+1},k_{m^{'}},{a_{k_{m^{'}}}^{\mathrm{ED}}}^* \right) = \\
				\underset{k^{'}\in \mathcal{A} ^{\mathrm{SW}}}{\max}Q_{t+1}\left( H_{t+1},k^{'},{a_{k^{'}}^{\mathrm{ED}}}^* \right) ,\forall H_{t+1}\in \mathcal{H} .
			\end{split}
		\end{equation}
	    Next, we can prove 
	    \begin{equation}
	    	\label{nextQt+1}
	    	\begin{split}
	    		&\left. \sum_{H_{t+1}\in \mathcal{H}}^{}{\mathrm{Pr}\left( H_{t+1}|H_t,k_1,{a_{k_1}^{\mathrm{ED}}}^* \right) \cdot \underset{k^{'}\in \mathcal{A} ^{\mathrm{SW}}}{\max}Q_t\left( H_{t+1},k^{'},{a_{k^{'}}^{\mathrm{ED}}}^* \right)} \right. \\
	    		&=\left. \sum_{H_{t+1}\in \mathcal{H}}^{}{\mathrm{Pr}\left( H_{t+1}|H_t,k_2,{a_{k_2}^{\mathrm{ED}}}^* \right) \cdot \underset{k^{'}\in \mathcal{A} ^{\mathrm{SW}}}{\max}Q_t\left( H_{t+1},k^{'},{a_{k^{'}}^{\mathrm{ED}}}^* \right)} \right. , \\
	    		&\forall k_1,k_2\in \mathcal{A} _{m}^{\mathrm{SW}}.
	    	\end{split}
	    \end{equation}
    This is because the history transition probability $\mathrm{Pr}\left( H_{t+1}|H_t,A_t \right)$ consists of four parts according to \eqref{prHt}, where the transition probabilities of load demands and PV output powers, i.e., $\mathrm{Pr}\left( P_{t}^{\mathrm{L}}|P_{t-\tau}^{\mathrm{L}},\dots,P_{t-1}^{\mathrm{L}} \right)$ and $\mathrm{Pr}\left( P_{t}^{\mathrm{PV}}|P_{t-\tau}^{\mathrm{PV}},\dots,P_{t-1}^{\mathrm{PV}} \right)$, do not depend on the switching action $k$. Moreover, all the switching actions $\forall k\in \mathcal{A} _{m}^{\mathrm{SW}}$ lead to the same SoC state $E_{t+1}$ according to \eqref{eq4}, \eqref{delta}, \eqref{ut} and \eqref{eq10}. Finally, the next switching state $B_{t+1}$ are different for $\forall k\in \mathcal{A} _{m}^{\mathrm{SW}}$, but the number of ON DGs at the next time step $t+1$ are the same, i.e., $m$. Therefore, \eqref{nextQt+1} is proved.

	 Finally, we can prove \eqref{thmeq} at each time step $t$ based on mathematical induction by considering the following Bellman equations in \eqref{bellman}, i.e.,
 \begin{align}
 	\label{iteration}
 	\underset{m\in \hat{\mathcal{A}}^{\mathrm{SW}}}{\max}Q_t\left( H_t,k_m,{a_{k_m}^{\mathrm{ED}}}^* \right) \notag \\
 	\overset{\left( a \right)}{=}\underset{m\in \hat{\mathcal{A}}^{\mathrm{SW}}}{\max}\left. \left\{ \underset{S_t|H_t}{\mathbb{E}}\left[r(S_t,k_m,{a_{k_m}^{\mathrm{ED}}}^*)\right] \right.\right.\notag \\
 	+\left. \left.\underset{H_{t+1}|H_t,k_m,{a_{k_m}^{\mathrm{ED}}}^*}{\mathbb{E}}\left[ \gamma \cdot \underset{m'\in \hat{\mathcal{A}}^{\mathrm{SW}}}{\max}Q_t\left( H_{t+1},k_{m'},{a_{k_{m'}}^{\mathrm{ED}}}^* \right) \right] \right. \right\} \notag\\
 	\overset{\left( b \right)}{=}\underset{m\in \hat{\mathcal{A}}^{\mathrm{SW}}}{\max}\left\{ \underset{S_t|H_t}{\mathbb{E}}\left[r(S_t,k_m,{a_{k_m}^{\mathrm{ED}}}^*)\right] \right.\notag\\
 	+\left.\underset{H_{t+1}|H_t,k_m,{a_{k_m}^{\mathrm{ED}}}^*}{\mathbb{E}}\left[ \gamma \cdot \underset{k'\in \mathcal{A} ^{\mathrm{SW}}}{\max}Q_t\left( H_{t+1},k',{a_{k'}^{\mathrm{ED}}}^* \right)  \right] \right\} \notag\\
 	\overset{\left( c \right)}{=}\underset{m\in \hat{\mathcal{A}}^{\mathrm{SW}}}{\max}\left. \underset{k\in \mathcal{A} _{m}^{\mathrm{SW}}}{\max}\left\{ \underset{S_t|H_t}{\mathbb{E}}\left[r(S_t,k,{a_{k}^{\mathrm{ED}}}^*)\right] \right.\right.\notag\\
 	+\left. \underset{H_{t+1}|H_t,k,{a_{k}^{\mathrm{ED}}}^*}{\mathbb{E}}\left[ \gamma \cdot \underset{k'\in \mathcal{A} ^{\mathrm{SW}}}{\max}Q_t\left( H_{t+1},k',{a_{k'}^{\mathrm{ED}}}^* \right)  \right] \right\} \notag\\
 	\overset{\left( d \right)}{=}\underset{k\in \mathcal{A} ^{\mathrm{SW}}}{\max}\left\{ \underset{S_t|H_t}{\mathbb{E}}\left[r(S_t,k,{a_{k}^{\mathrm{ED}}}^*)\right] \right. \notag\\
 	+\left.\underset{H_{t+1}|H_t,k,{a_{k}^{\mathrm{ED}}}^*}{\mathbb{E}}\left[ \gamma \cdot \underset{k'\in \mathcal{A} ^{\mathrm{SW}}}{\max}Q_t\left( H_{t+1},k',{a_{k'}^{\mathrm{ED}}}^* \right)  \right] \right\} \notag\\
 	\overset{\left( e \right)}{=}\underset{k\in \mathcal{A} ^{\mathrm{SW}}}{\max}Q_t\left( H_t,k,{a_{k}^{\mathrm{ED}}}^* \right) .
 \end{align}
 \noindent where (a) and (e) are due to the definition of Bellman equation; (b) follows by \eqref{thmeqt+1}; (c) is due to \eqref{nextQt+1} and the definition of expectation; (d) follows by \eqref{cup}. Therefore, \eqref{thmeq} in Theorem 1 is proved according to \eqref{iteration}.
	\end{proof}
\end{appendices}
\bibliography{author}
\bibliographystyle{IEEEtran}

\end{document}